\documentclass[english]{article}
\usepackage[T1]{fontenc}
\usepackage[latin9]{inputenc}
\usepackage{geometry}
\usepackage{float}
\usepackage{amsmath}
\usepackage{amssymb}
\usepackage{graphicx}
\usepackage{esint}
\usepackage[authoryear]{natbib}

\makeatletter

\providecommand{\tabularnewline}{\\}
\floatstyle{ruled}
\newfloat{algorithm}{tbp}{loa}
\providecommand{\algorithmname}{Algorithm}
\floatname{algorithm}{\protect\algorithmname}

\newcommand{\lyxaddress}[1]{
\par {\raggedright #1
\vspace{1.4em}
\noindent\par}
}

\usepackage{times}

\usepackage{graphicx} 
\usepackage{subfigure}

\usepackage{natbib}

\usepackage{algorithm}
\usepackage{algorithmic}

\usepackage{hyperref}

\makeatother

\usepackage{babel}
\begin{document}

\title{Mean Field Bayes Backpropagation: scalable training of multilayer
neural networks with binary weights}

\author{Daniel Soudry{*}, Ron Meir}

\maketitle

\lyxaddress{Department of Electrical Engineering, Technion 32000, Haifa, Israel\\
{*}Corresponding author: daniel.soudry@gmail.com}
\begin{abstract}
Significant success has been reported recently ucsing deep neural
networks for classification. Such large networks can be computationally
intensive, even after training is over. Implementing these trained
networks in hardware chips with a limited precision of synaptic weights
may improve their speed and energy efficiency by several orders of
magnitude, thus enabling their integration into small and low-power
electronic devices. With this motivation, we develop a computationally
efficient learning algorithm for multilayer neural networks with binary
weights, assuming all the hidden neurons have a fan-out of one. This
algorithm, derived within a Bayesian probabilistic online setting,
is shown to work well for both synthetic and real-world problems,
performing comparably to algorithms with real-valued weights, while
retaining computational tractability.

\end{abstract}

\section{Introduction}

Recently, Multilayer%
\footnote{\emph{i.e.}, having more than a single layer of\emph{ }adjustable
weights.%
} Neural Networks (MNNs) with deep architecture have achieved state-of-the-art
performance in various machine learning tasks \citep{Deng2012,Krizhevsky2012,Dahl2012}
- even when only supervised on-line gradient descent algorithms are
used \citep{Dean2012,Ciresan2012a,Ciresan2012b}. However, it is not
clear what would be the best choice of a training algorithm for such
networks \citep{Le2011a}. Within a Bayesian setting, given some prior
distribution and error function, we can choose an on-line optimal
estimate based on updating the posterior distribution. Such a ``Bayesian''
approach has relatively transparent assumptions and, furthermore,
gives estimates of learning uncertainty and allows model averaging.
However, an exact Bayesian computation is generally intractable. Bayesian
approaches, based on Monte Carlo simulations, were previously used
\citep{Mackay1992,Neal1995}, but these are not generally scalable
with the size of the network and training data. Other approximate
Bayesian approaches were suggested \citep{Opper1996,Winther1997,Opper1998,Sollaa},
but only for Single-layer%
\footnote{\emph{i.e.}, having only a single layer of\emph{ }adjustable weights.%
} Neural Networks (SNN). To the best of our knowledge, it is still
unknown whether such methods could be generalized to multilayer networks. 

Another advantage for a Bayesian approach is that it can be used even
when gradients do not exist. For example, it could be very useful
if weights are restricted to assume only binary values (\emph{e.g.},
$\pm1$). This may allow a dense, fast and energetically efficient
hardware implementation of MNNs (\emph{e.g.}, with the chip in \citep{Karakiewicz2012},
which can perform $10^{12}$ operations per second with $1\mathrm{mW}$
power efficiency). Limiting the weights to binary values only mildly
reduces the (linear) computational capacity of a MNN (at most, by
a logarithmic factor \citep{Ji1998}). However, learning in a Binary
MNN (BMNN - a MNN with binary weights) is much harder than learning
in a Real-valued MNN (RMNN - a MNN with real valued weights). For
example, if the weights of a single neuron are restricted to binary
values, the computational complexity of learning a linearly separable
set of patterns becomes NP-hard (instead of P) in the dimension of
the input \citep{Fang1996}. In spite of this, it is possible to train
single binary neurons in a \emph{typical} linear time \citep{Fang1996}.
Interestingly, the most efficient methods developed for training single
binary neurons use approximate Bayes approaches, either explicitly
\citep{Sollaa,Ribeiro2011} or implicitly \citep{Braunstein,Baldassi2007a}. 

However, as far as we are aware, it remains an open question whether
a BMNN can be trained efficiently by such Bayesian methods, or by
any other method. Standard RMNNs are commonly trained in supervised
mode using the Backpropagation algorithm \citep{Lecun1998}. However,
it is unsuitable if the weight values are binary (crude discretization
of the weights is usually quite destructive \citep{Moerland1997}).
Other, non-Bayesian  methods were suggested in the 90's (\emph{e.g.},
\citep{Saad1990,Battiti1995,Mayoraz1996}) for small BMNNs, but it
is not clear whether these approaches are scalable. 

In this work we derive a Mean Field Bayes Backpropagation (MFB-BackProp)
algorithm for learning synaptic weights in BMNNs where each hidden
neuron has only a single outgoing connection (however, the input layer
can be fully connected to the first neuronal layer, see Fig. \ref{fig:The-architecture-of-network}).
This algorithm has linear computational complexity in the number of
weights, similarly to standard Backpropagation. Also, it is parameter-free
except for the initial conditions (``prior''). The algorithm implements
a Bayes update to the weights, using two approximations (as was done
in \citep{Sollaa,Ribeiro2011} for SNNs): \emph{(1)} a mean-field
approximation - the posterior probability of the synaptic weights
is approximated by the product of its marginals at each time step,
\emph{(2)} the fan-in of all neurons is large (so their inputs are
approximately Gaussian). Despite these approximations, we demonstrate
numerically that the algorithm works well. First, we demonstrated
its effectiveness in a synthetic teacher-student scenario where the
outputs are generated from a network with a known architecture. The
network performed well even though the assumption of large fan-in
did not hold. We then tested it for a large BMNN ($\sim10^{6}$ weights)
on the standard MNIST task, achieving an error rate comparable with
the RMNN of similar size. Therefore, as far as we are aware, it is
the first scalable algorithm for BMNNs and the first scalable Bayesian
algorithm for MNNs in general (\emph{i.e.}, it does not require Monte
Carlo simulations, as in \citep{Mackay1992,Neal1995}). Hopefully,
the methods developed here could lead to scalable hardware implementations,
as well as efficient Bayesian-based learning in MNNs.

\section{Preliminaries\label{sec:Preliminaries}}

\begin{figure}
\begin{centering}
\includegraphics[width=0.8\columnwidth]{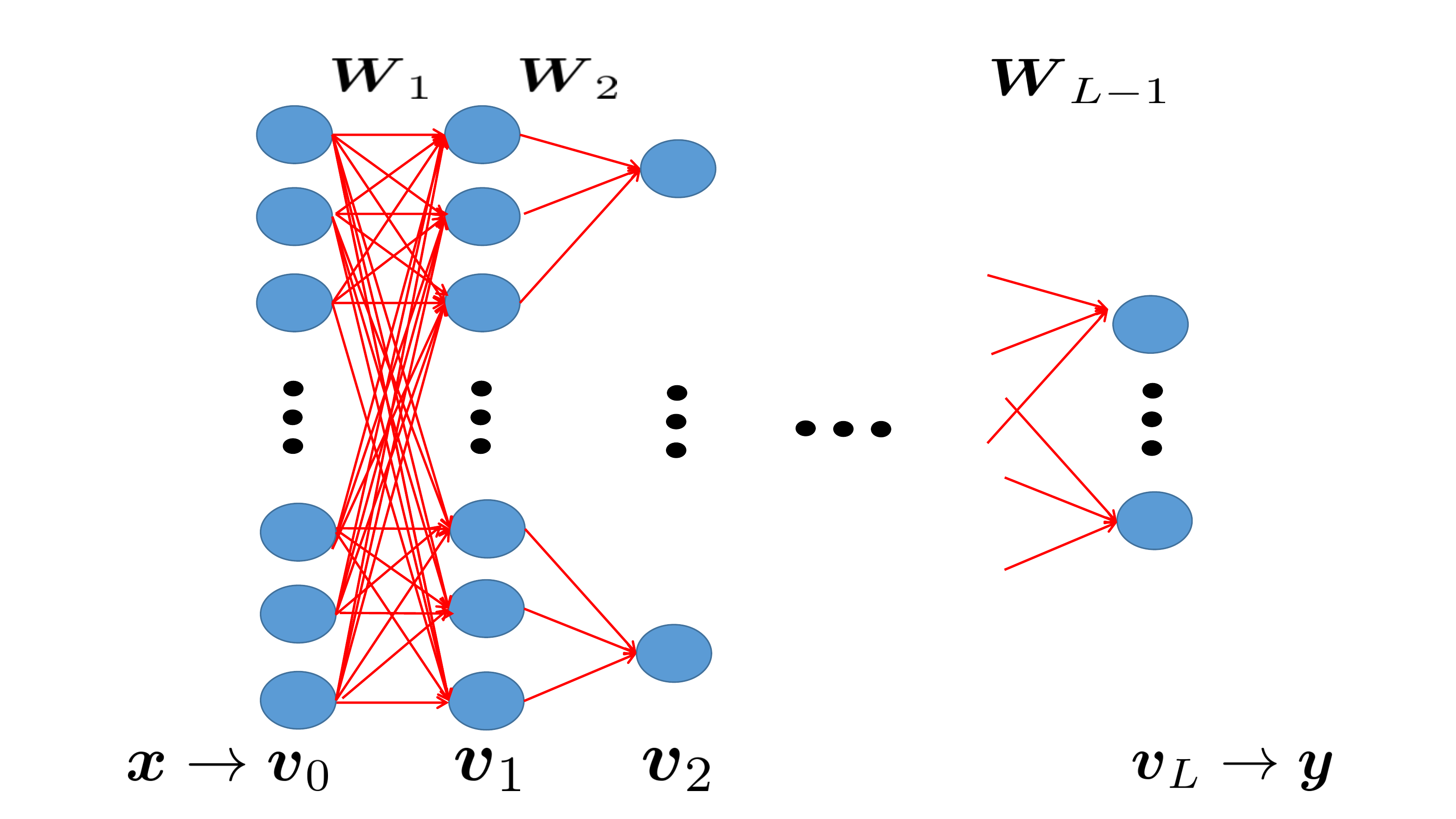}
\par\end{centering}

\caption{The convergent architecture of the network - all hidden neurons have
fan-out 1, while input layer is fully connected.\label{fig:The-architecture-of-network}}
\end{figure}

\paragraph{Notation}

We denote by $P\left(x\right)$ the probability distribution (in the
discrete case) or density (in the continuous case) of a random variable
$X$, $P\left(x|y\right)=P\left(x,y\right)/P\left(y\right)$,$\left\langle x\right\rangle =\int xP\left(x\right)dx$,
$\left\langle x|y\right\rangle =\int xP\left(x|y\right)dx$, $\mathrm{Cov}\left(x,y\right)=\left\langle xy\right\rangle -\left\langle x\right\rangle \left\langle y\right\rangle $
and $\mathrm{Var}\left(x\right)=\mathrm{Cov}\left(x,x\right)$. Integration
is exchanged with summation in the discrete case. Furthermore, we
make use of the following functions: \emph{(1)} $\theta\left(x\right)$,
the Heaviside function (\emph{i.e.} $\theta\left(x\right)=1$ for
$x>0$ and zero otherwise), \emph{(2)} $\delta_{ij}$, the Kronecker
delta function (\emph{i.e.} $\delta_{ij}=1$ if $i=j$ and zero otherwise).
Also,\emph{ }If $\mathbf{x}\sim\mathcal{N}\left(\boldsymbol{\mu},\boldsymbol{\Sigma}\right)$
then it is Gaussian with mean $\boldsymbol{\mu}$ and covariance matrix
$\boldsymbol{\Sigma}$, and we denote its density by $\mathcal{N}\left(\mathbf{x}|\boldsymbol{\mu},\boldsymbol{\Sigma}\right)$.
Finally, $\Phi\left(x\right)=\int_{-\infty}^{x}\mathcal{N}\left(x|0,1\right)dx$,
is the Gaussian cumulative distribution function.

\paragraph{Model}

We consider a BMNN with $L$ layers of synaptic weight matrices $\mathcal{W}=\left\{ \mathbf{W}_{l}\in\left\{ -1,1\right\} ^{V_{l}\times V_{l-1}}\right\} _{l=1}^{L}$
connecting neuronal layers sequentially, with $V_{l}$ being the width
of the $l$-th layer. We denote the outputs of the layers by $\left\{ \mathbf{v}_{l}\right\} _{l=0}^{L}$,
where $\mathbf{v}_{0}$ is the input layer, $\left\{ \mathbf{v}_{l}=\mathrm{sign}\left(\mathbf{W}_{l}\mathbf{v}_{l-1}\right)\in\left\{ -1,1\right\} ^{V_{l}}\right\} _{l=1}^{L}$
are the hidden layers (with the $\mathrm{sign}\mathrm{\left(\cdot\right)}$
function operating component-wise) and the $\mathbf{v}_{L}$ is the
output layer. The output of the network is therefore 
\begin{equation}
\mathbf{v}_{L}=f\left(\mathbf{v}_{0},\mathcal{W}\right)=\mathrm{sign}\left(\mathbf{W}_{L}\mathrm{sign}\left(\mathbf{W}_{L-1}\mathrm{sign}\left(\cdots\mathbf{W}_{1}\mathbf{v}_{0}\right)\right)\right)\,.\label{eq: BMNN output}
\end{equation}
Also, we denote $W_{ij,l}=\left(\mathbf{W}_{l}\right)_{ij}$ and $v_{i,l}=\left(\mathbf{v}_{l}\right)_{i}$.
Furthermore, we assume the network has a \emph{converging architecture},
\emph{i.e.}, each hidden neuron has only a single outgoing connection,
namely, fan-out$=1$. However, the input layer can be fully connected
to the first neuronal layer (Fig. \ref{fig:The-architecture-of-network}).
We set $K\left(i,l\right)$ to be the set of indices of neurons in
the $\left(l-1\right)$-th layer connected to the $i$-th neuron in
the $l$-th layer. For simplicity we assume that $K_{l}=\left|K\left(i,l\right)\right|$
are constant for all $i$. Finally, we denote $i^{\prime}$ to be
the index of the neuron in the $\left(l+1\right)$-th layer receiving
input from the $i$-th neuron in the $l$-th layer.

\paragraph{Task}

We examine a standard supervised learning classification task, in
which we are given sequentially labeled data points $D_{N}=\left\{ \mathbf{x}^{\left(n\right)},\mathbf{y}^{\left(n\right)}\right\} _{n=1}^{N}$,
where $\mathbf{x}^{\left(n\right)}\in\mathbb{R}^{V_{0}}$ is the data
point, $\mathbf{y}^{\left(n\right)}\in\left\{ -1,1\right\} ^{V_{L}}$
is the label and $n$ is the sample index (for brevity, we will sometimes
suppress the sample index, where it is clear from the context). As
common for supervised learning with MNNs, we assume that $\forall n$
the relation $\mathbf{y}^{\left(n\right)}=f\left(\mathbf{x}^{\left(n\right)},\mathcal{W}\right)$
can be represented by a BMNN with known architecture (the `hypothesis
class') and unknown weights $\mathcal{W}$ (\emph{i.e}, according
to Eq.~\ref{eq: BMNN output} with $\mathbf{v}_{0}=\mathbf{x}^{\left(n\right)}$
and \textbf{$\mathbf{v}_{L}=\mathbf{y}^{\left(n\right)}$}). Our goal
is to estimate the weights $\mathcal{W}$.

\section{Theory - online Bayesian learning in BMNNs}

We approach this task from a Bayesian framework, where we estimate
the Maximum A Posteriori (MAP) configuration of weights 
\begin{equation}
\mathcal{W}^{*}=\mathrm{argmax}_{\mathcal{W}}P\left(\mathcal{W}|D_{N}\right)\,,\label{eq: MAP}
\end{equation}
with $P\left(\mathcal{W}|D_{N}\right)$ being the probability for
each configuration of the weights $\mathcal{W}$, given the data.
We do this in an online setting. Starting with some prior distribution
on the weights - $P\left(\mathcal{W}|D_{0}\right)$, we update the
value of $P\left(\mathcal{W}|D_{n}\right)$ after the $n$-th sample
is received, according to Bayes rule:
\begin{equation}
P\left(\mathcal{W}|D_{n}\right)\propto P\left(\mathbf{y}^{\left(n\right)}|\mathbf{x}^{\left(n\right)},\mathcal{W}\right)P\left(\mathcal{W}|D_{n-1}\right)\,,\label{eq: bayes update}
\end{equation}
for $n=1,\dots,N$. Note that the BMNN is \emph{deterministic}, so
\begin{equation}
P\left(\mathbf{y}^{\left(n\right)}|\mathbf{x}^{\left(n\right)},\mathcal{W}\right)=\begin{cases}
1 & ,\,\mathrm{if}\,\,\mathbf{y}^{\left(n\right)}=f\left(\mathbf{x}^{\left(n\right)},\mathcal{W}\right)\\
0 & ,\,\mathrm{otherwise}
\end{cases}\,.\label{eq: P(y|x,W) indicator}
\end{equation}
Therefore, the Bayes update in Eq.~\ref{eq: bayes update} simply
makes sure that $P\left(\mathcal{W}|D_{n}\right)=0$ in any ``illegal''
configuration (\emph{i.e.}, any $\mathcal{W}^{0}$ such that $\exists k\leq n:\,\mathbf{y}^{\left(k\right)}\neq f\left(\mathbf{x}^{\left(k\right)},\mathcal{W}^{0}\right)$).
Unfortunately, this update is generally intractable for large networks,
mainly because we need to store and update an exponential number of
values for $P\left(\mathcal{W}|D_{n}\right)$. Therefore, some approximation
is required.

\subsection{Approximation 1: mean-field}

Instead of storing $P\left(\mathcal{W}|D_{n}\right)$, we will store
its factorized (`mean-field') approximation $\hat{P}\left(\mathcal{W}|D_{n}\right)$,
for which
\begin{equation}
\hat{P}\left(\mathcal{W}|D_{n}\right)=\prod_{i,j,l}\hat{P}\left(W_{ij,l}|D_{n}\right)\,,\label{eq: factorization assumption}
\end{equation}
where each factor must be normalized. Notably, it is easy to find
the MAP estimate (Eq.~~\ref{eq: MAP}) under this factorized approximation
$\forall i,j,l$
\begin{equation}
W_{ij,l}^{*}=\mathrm{argmax}_{W_{ij,l}\in\left\{ -1,1\right\} }\hat{P}\left(W_{ij,l}|D_{N}\right)\,.\label{eq: MAP-marginal}
\end{equation}
The factors $\hat{P}\left(W_{ij,l}|D_{n}\right)$ can be found using
a standard variational approach \citet{Bishop2006,Sollaa}. For each
$n$, we first perform the Bayes update in Eq.~\ref{eq: bayes update}
with $\hat{P}\left(\mathcal{W}|D_{n-1}\right)$ instead of $P\left(\mathcal{W}|D_{n-1}\right)$.
Then, we project the resulting posterior onto the family of distributions
factorized as in Eq.~\ref{eq: factorization assumption}, by minimizing
the reverse Kullback-Leibler divergence. A straightforward calculation
shows that the optimal factor is just a marginal of the posterior
(see supplemental material, section \ref{sec:The-mean-field-approximation}).
Performing this marginalization on the Bayes update and re-arranging
terms, we obtain $\forall i,j,l$
\begin{align}
\hat{P}\left(W_{ij,l}|D_{n}\right) & \propto\hat{P}\left(\mathbf{y}^{\left(n\right)}|\mathbf{x}^{\left(n\right)},W_{ij,l},D_{n-1}\right)\hat{P}\left(W_{ij,l}|D_{n-1}\right)\,,\label{eq: bayes update-1 - L}
\end{align}
 where
\begin{eqnarray}
\hat{P}\left(\mathbf{y}^{\left(n\right)}|\mathbf{x}^{\left(n\right)},W_{ij,l},D_{n-1}\right) & = & \sum_{\mathbf{\mathcal{W}}^{\prime}:W_{ij,l}^{\prime}=W_{ij,l}}P\left(\mathbf{y}^{\left(n\right)}|\mathbf{x}^{\left(n\right)},\mathcal{W}^{\prime}\right)\prod_{\left\{ k,r,m\right\} \neq\left\{ i,j,l\right\} }\hat{P}\left(W_{kr,m}^{\prime}|D_{n-1}\right)\,.\label{eq: P(y|x,W,D) - L}
\end{eqnarray}
 Thus we can directly update the factor $\hat{P}\left(W_{ij,l}|D_{n}\right)$
in a single step. However, the last equation is still problematic,
since it contains a generally intractable summation over an exponential
number of values, and therefore requires simplification. For brevity,
from now on we replace any $\hat{P}$ with $P$, in a slight abuse
of notation (keeping in mind that the distributions are approximated).

\subsection{Simplifying the update rule}

In order to be able to use the update rule in Eq.~\ref{eq: bayes update-1 - L},
we wish to calculate $P\left(\mathbf{y}^{\left(n\right)}|\mathbf{x}^{\left(n\right)},W_{ij,l},D_{n-1}\right)$
using Eq.~\ref{eq: P(y|x,W,D) - L}. For brevity, we suppress the
$\left(n\right)$ index and the dependence on $D_{n-1}$ and \textbf{$\mathbf{x}$}
\begin{eqnarray}
P\left(\mathbf{y}|W_{ij,l}\right) & = & \!\!\!\!\!\sum_{\mathbf{\mathcal{W}}^{\prime}:W_{ij,l}^{\prime}=W_{ij,l}}\!\!\!\!\!\!\!\!\! P\left(\mathbf{y}|\mathcal{W}^{\prime}\right)\!\!\!\!\!\!\!\!\!\prod_{\left\{ k,r,m\right\} \neq\left\{ i,j,l\right\} }\!\!\!\!\!\!\!\!\! P\left(W_{kr,m}^{\prime}\right)\label{eq:P(y|x,Wij,l) - L}
\end{eqnarray}
Since, by assumption, $P\left(\mathbf{y}|\mathcal{W}^{\prime}\right)$
comes from a feed-forward BMNN with input $\mathbf{x}$, we can re-write
Eq.~\ref{eq: P(y|x,W) indicator} as
\begin{eqnarray}
P\left(\mathbf{y}|\mathcal{W}^{\prime}\right) & = & \sum_{\mathbf{v}_{1},\dots,\mathbf{v}_{L-1}}\prod_{m=1}^{L}\prod_{k=1}^{V_{m}}\theta\left(v_{k,m}\!\!\!\!\!\sum_{r\in K\left(k,m\right)}\!\!\!\!\! v_{r,m-1}W_{kr,m}^{\prime}\right),\label{eq: p(y|x,W') - L}
\end{eqnarray}
where $\mathbf{v}_{L}=\mathbf{y}\mbox{ and }\mathbf{v}_{0}=\mathbf{x}$,
and we recall that $\theta\left(x\right)=1$ for $x>0$ and zero otherwise
(consequently, only a single term in the summation is non-zero). Substituting
Eq.~\ref{eq: p(y|x,W') - L}  into Eq.~\ref{eq:P(y|x,Wij,l) - L},
allows us to perform the summations in a more convenient way - layer
by layer. To do this, we define
\begin{eqnarray}
P\left(\mathbf{v}_{m}|\mathbf{v}_{m-1}\right) & = & \!\sum_{\mathbf{W}_{m}^{\prime}}\!\prod_{k=1}^{V_{m}}\!\left[\!\theta\left(v_{k,m}\!\!\!\!\!\!\!\sum_{r\in K\left(k,m\right)}\!\!\!\!\!\!\!\! v_{r,m-1}W_{kr,m}^{\prime}\!\!\right)\!\!\!\prod_{r\in K\left(k,m\right)}\!\!\!\!\!\! P\left(W_{kr,m}^{\prime}\right)\right]\label{eq:P(v|v) - L}
\end{eqnarray}
and $P\left(\mathbf{v}_{l}|\mathbf{v}_{l-1},W_{ij,l}\right)$, which
is defined identically to $P\left(\mathbf{v}_{l}|\mathbf{v}_{l-1}\right)$,
except the summation is done over all configurations in which $W_{ij,l}$
is fixed (\emph{i.e.}, $\mathbf{W}_{l}^{\prime}:W_{ij,l}^{\prime}=W_{ij,l}$)
and we set $P\left(W_{ij,l}\right)=1$. Now we can write recursively
\begin{eqnarray}
\!\!\!\!\!\!\!\!\!\! P\left(\mathbf{v}_{1}\right)\!\!\!\!\! & = & \!\!\!\! P\left(\mathbf{v}_{1}|\mathbf{v}_{0}=\mathbf{x}\right)\label{eq: P(v1)}\\
\!\!\!\!\!\!\!\!\!\! P\left(\mathbf{v}_{m}\right)\!\!\!\!\! & = & \!\!\!\!\sum_{\mathbf{v}_{m-1}}P\left(\mathbf{v}_{m}|\mathbf{v}_{m-1}\right)P\left(\mathbf{v}_{m-1}\right)\label{eq:P(v|x)}\\
\!\!\!\!\!\!\!\!\!\! P\left(\mathbf{v}_{l}|W_{ij,l}\right)\!\!\!\!\! & = & \!\!\!\!\sum_{\mathbf{v}_{l-1}}P\left(\mathbf{v}_{l}|\mathbf{v}_{l-1},W_{ij,l}\right)P\left(\mathbf{v}_{l-1}\right)\label{eq:P(vl | W)}\\
\!\!\!\!\!\!\!\!\!\! P\left(\mathbf{v}_{m}|W_{ij,l}\right)\!\!\!\!\! & = & \!\!\!\!\sum_{\mathbf{v}_{m-1}}P\left(\mathbf{v}_{m}|\mathbf{v}_{m-1}\right)P\left(\mathbf{v}_{m-1}|W_{ij,l}\right)\label{eq:P(vm| W)}
\end{eqnarray}
where Eq.~\ref{eq:P(v|x)} is $\forall m\in\left\{ 2,..,l-1\right\} $,
and Eq.~\ref{eq:P(vm| W)} is $\forall m\in\left\{ l+1,l+2,..,L\right\} $.
Thus we obtain the result of Eq.~\ref{eq:P(y|x,Wij,l) - L}, through
$P\left(\mathbf{y}|W_{ij,l}\right)=P\left(\mathbf{v}_{L}=\mathbf{y}|W_{ij,l}\right)$.
However, this problem is still generally intractable, since all of
the above summations (Eqs.~\ref{eq:P(v|v) - L}-\ref{eq:P(vm| W)})
are still over an exponential number of values. Therefore, we need
to make one additional approximation.

\subsection{Approximation 2: large fan-in}

In order to simplify the above summations (Eqs.~\ref{eq:P(v|v) - L}-\ref{eq:P(vm| W)}),
we assume that the fan-in of all of the connections is ``large''.
In the rest of this section, we summarize the results obtained using
this approximation (see details in supplementary material, section \ref{sec:Forward-propagation-of}).
Importantly, if $K_{l}\rightarrow\infty$ then we can use the Central
Limit Theorem (CLT) and say that the normalized input to each neuronal
layer, is distributed according to a Gaussian distribution
\begin{equation}
\forall m:\,\,\mathbf{u}_{m}=\mathbf{W}_{m}\mathbf{v}_{m-1}/\sqrt{K_{m}}\sim\mathcal{N}\left(\boldsymbol{\mu}_{m},\boldsymbol{\Sigma}_{m}\right)\,,\label{eq: CLT}
\end{equation}
Since $K_{l}$ is actually finite, this is only an approximation -
though a quite common and effective one (\emph{e.g.}, \citet{Neal1995,Ribeiro2011}).
Using the approximation in Eq.~\ref{eq: CLT} and Eqs.~\ref{eq: P(v1)}-\ref{eq:P(v|x)},
we can now calculate the distribution of $\mathbf{u}_{m}$ sequentially
for all the layers $m\in\left\{ 1,\dots,l-1\right\} $, by deriving
$\boldsymbol{\mu}_{m}$ and $\boldsymbol{\Sigma}_{m}$ for each layer.
Additionally, due to the ``converging'' architecture of the network
(\emph{i.e.}, fan-out=1) it is easy to show that $\Sigma_{kr,m}=\delta_{kr}\sigma_{k,m}^{2}$.
Therefore, since $\mathbf{v}_{m}=\mathrm{sign}\left(\mathbf{u}_{m}\right)$,
we can use simple Gaussian integrals on $\mathbf{u}_{m}$ and obtain
$\forall m\geq1$ 
\[
P\left(\mathbf{v}_{m}\right)=\prod_{k}P\left(v_{k,m}\right)=\prod_{k}\Phi\left(v_{k,m}\mu_{k,m}/\sigma_{k,m}\right)\,,
\]
as the solution of Eqs.~\ref{eq: P(v1)}-\ref{eq:P(v|x)}. This
immediately gives 
\begin{equation}
\left\langle v_{k,m}\right\rangle =v_{k,0}\delta_{0m}+\left(1-\delta_{0m}\right)\left(2\Phi\left(\mu_{k,m}/\sigma_{k,m}\right)-1\right)\,.\label{eq: output}
\end{equation}
Note that $W_{kr,m}$ and $v_{k,m-1}$ are binary and independent
for a fixed $m$ (from Eqs.~\ref{eq: factorization assumption} and
\ref{eq: BMNN output}). Therefore, it is straightforward to derive
\begin{align}
\mu_{k,m} & =\frac{1}{\sqrt{K_{m}}}\!\sum_{r\in K\left(k,m\right)}\!\!\!\!\!\!\left\langle W_{kr,m}\right\rangle \left\langle v_{r,m-1}\right\rangle \label{eq: mu}\\
\sigma_{k,m}^{2} & =\frac{1}{K_{m}}\!\sum_{r\in K\left(k,m\right)}\!\!\!\!\!\!\left(\left\langle \left|v_{r,m-1}\right|\right\rangle ^{2}-\left\langle v_{r,m-1}\right\rangle ^{2}\left\langle W_{kr,m}\right\rangle ^{2}\right),\nonumber 
\end{align}
where we note that $\left\langle \left|v_{r,m}\right|\right\rangle =1+\delta_{m0}\left(v_{r,0}-1\right)$. 

Next, we repeat similar derivations for $m\in\left\{ l,\dots,L\right\} $
(Eqs.~\ref{eq:P(vl | W)} and \ref{eq:P(vm| W)}). Note that for
layer $l$, $W_{ij,l}$ is fixed (not a random variable), so we need
slightly modified versions of the mean $\mu_{k,m}$ and variance $\sigma_{k,m}^{2}$
in which $W_{ij,m}$ is ``disconnected''. Thus, we define $\mu_{i\left(j\right),l}$
and $\sigma_{i\left(j\right),l}^{2}$, which are identical to $\mu_{k,m}$
and $\sigma_{k,m}^{2}$, except the summation is done over $K\left(i,l\right)\backslash j$,
instead of $K\left(i,l\right)$. Importantly, if we know $\mathbf{x}$
and $\left\langle W_{kr,1}\right\rangle =2P\left(W_{kr,1}=1\right)-1$,
all these quantities can be calculated together in a sequential ``forward
pass'' for $m=1,2,...,L$. At the end of this forward pass we will
be able to find $P\left(\mathbf{y}|W_{ij,l}\right)=P\left(\mathbf{v}_{L}=\mathbf{y}|W_{ij,l}\right)$.
However, it is more convenient to derive instead the log-likelihood
ratio (see the supplementary material, sections \ref{sec:The-log-likelihood-ratio}
and \ref{sec:Summary})
\[
R_{ij,l}=\ln\frac{P\left(\mathbf{y}|W_{ij,l}=1\right)}{P\left(\mathbf{y}|W_{ij,l}=-1\right)}\,.
\]
This quantity is useful, since, from Eq.~\ref{eq: bayes update-1 - L},
it uniquely determines the Bayes updates of the posterior through
\begin{equation}
\ln\frac{P\left(W_{ij,l}=1|D_{n}\right)}{P\left(W_{ij,l}=-1|D_{n}\right)}=\ln\frac{P\left(W_{ij,l}=1|D_{n-1}\right)}{P\left(W_{ij,l}=-1|D_{n-1}\right)}+R_{ij,l}^{\left(n\right)}\,.\label{eq: bayes update-2}
\end{equation}
To obtain $R_{ij,l}$ we define
\begin{equation}
G_{ij,l}=\frac{2}{\sqrt{K_{l}}}\frac{\mathcal{N}\left(0|\mu_{i\left(j\right),l},\sigma_{i\left(j\right),l}^{2}\right)}{1-\delta_{Ll}+\Phi\left(y_{i}\mu_{i\left(j\right),L}/\sigma_{i\left(j\right),L}\right)\delta_{Ll}}\label{eq: G}
\end{equation}
and a backward propagated `delta' quantity, defined recursively so
that $\Delta_{ij,L+1}=y_{j}$, and for $l=L,L-1,\dots2$,
\begin{equation}
\Delta_{ij,l}=\Delta_{i^{\prime}i,l+1}\left\langle W_{ij,l}\right\rangle \tanh\left[G_{ij,l}\right]\,,\label{eq:Delta ij.l}
\end{equation}
where we recall that $i^{\prime}$ is the index of the neuron in the
$\left(m+1\right)$-th layer receiving input from the $i$-th neuron
in the $m$-th layer. Using the above quantities, we finally arrive
at the log-likelihood ratio
\begin{equation}
R_{ij,l}=\begin{cases}
\Delta_{i^{\prime}i,l+1}\tanh\left[G_{ij,l}x_{j}\right] & ,\,\mathrm{if}\,\, l=1\\
\Delta_{i^{\prime}i,l+1}\tanh\left[G_{ij,l}\right]\left\langle v_{j,l-1}\right\rangle  & ,\,\mathrm{if}\,\, l>1
\end{cases}\,.\label{eq: loglikelihood}
\end{equation}

The derivation of these results is based on the assumption $K_{l}$
is ``large''. This is done so that: (1) we can use CLT in each layer
(2) we can assume that a flip of a single weight has a relatively
small affect on the output of the BMNN (\emph{i.e.}, $R_{ij,l}\ll1$),
and therefore we can use a first order Taylor approximation. However,
if $K_{l}$ is finite, both (1) and (2) can break down. Specifically,
this can occur if in a certain layer we have high ``certainty''
in our estimate (\emph{i.e.}, $\sigma_{i,l}^{\left(n\right)}\rightarrow0$,
so for all the weights $W_{kr,m}$ leading into $v_{i,l}$ for all
$m\leq l$ we have $P\left(W_{kr,m}=1|D_{n}\right)\approx0$ or $1$),
and then we receive a data point $\left\{ \mathbf{x}^{\left(n\right)},\mathbf{y}^{\left(n\right)}\right\} $
which is ``surprising'' (\emph{i.e.}, it does not conform with our
current estimate of these weights). In this case divergent inaccuracies
may occur since our assumptions break down. To prevent this scenario
we heuristically used a saturating function $\tanh\left(\cdot\right)$
in our derivations.

\section{Algorithm}

Now we can write down explicitly how $P\left(W_{ij,l}|D_{n}\right)$
changes, for every $W_{ij,l}$ . For convenience, we will parametrize
the distribution of $W_{ij,l}$ so that 
\[
P\left(W_{ij,l}|D_{n}\right)=e^{h_{ij,l}^{\left(n\right)}W_{ij,l}}/\left(e^{h_{ij,l}^{\left(n\right)}}+e^{-h_{ij,l}^{\left(n\right)}}\right)
\]
 and increment the parameter $h_{ij,l}^{\left(n\right)}$, according
to the Bayes-based update rule in Eq.~$\ref{eq: bayes update-2}$
\begin{equation}
h_{ij,l}^{\left(n\right)}=h_{ij,l}^{\left(n-1\right)}+\frac{1}{2}R_{ij,l}^{\left(n\right)},\,\label{eq: delta h}
\end{equation}
using $R_{ij,l}$, the log-likelihood ratio we obtained (Eq.~\ref{eq: loglikelihood}).
Additionally, we denote $\left(\mathbf{H}_{l}\right)_{ij}=h_{ij,l}$
, $\mathcal{H}=\left\{ \mathbf{H}_{l}\right\} _{l=1}^{L}$, $\nu_{k,l}=\left\langle v_{k,l-1}\right\rangle $
and note that $\left\langle W_{ij,l}\right\rangle =\tanh\left(h_{ij,l}\right)$
and $\mathrm{Var}\left(W_{ij,l}\right)=1-\tanh^{2}\left(h_{ij,l}\right)=\mathrm{sech}^{2}\left(h_{ij,l}\right)$.
Doing this entire calculation separately for each $h_{ij,l}$ is highly
inefficient - requiring about $O\left(\left|\mathcal{W}\right|^{2}\right)$
computation steps. We suggest the Mean Field Bayes Back Propagation
(MFB-BackProp) algorithm \ref{alg:The-Variational-Bayes} to do this
efficiently, in $O\left(\left|\mathcal{W}\right|\right)$ computation
steps, similarly to the original Backpropagation algorithm. The resulting
algorithm itself is rather similar to standard Backpropagation, as
we shall explain soon. After $\mathcal{H}$ is estimated using the
algorithm, the MAP estimate (Eq.~\ref{eq: MAP-marginal}) for the
BMNN is obtained by simple clipping 
\begin{equation}
W_{ij,l}^{*}=\mathrm{argmax}_{W_{ij,l}\in\left\{ -1,1\right\} }P\left(W_{ij,l}|D_{N}\right)=\mathrm{sign}\left(h_{ij,l}\right)\,.\label{eq: W MFB-BackProp}
\end{equation}
The output of this MAP BMNN is then given by Eq.~\ref{eq: BMNN output}.
However, the mean Bayes output given input $\mathbf{x}$ is
\[
\boldsymbol{\nu}_{L}=\left\langle \mathbf{v}_{L}\right\rangle =\sum_{\mathcal{W}}f\left(\mathbf{x},\mathcal{W}\right)P\left(\mathcal{W}|D_{N}\right)\,.
\]
Therefore, since $\mathbf{\mathbf{v}}_{L}$ is binary, the MAP \emph{output
}is simply
\begin{equation}
\hat{\mathbf{y}}_{p}=\mathrm{argmax}_{\mathbf{v}_{L}}P\left(\mathbf{v}_{L}\right)=\mathrm{sign}\left(\boldsymbol{\nu}_{L}\right)\,.\label{eq: y_p}
\end{equation}
Though only the MFB-BackProp output in Eq.~\ref{eq: BMNN output}
can be implemented in an actual binary circuit, this ``Probabilistic''
MFB-BackProp (PMFB-BackProp) output (Eq.~\ref{eq: y_p}) can be viewed
as an ensemble average of such circuits over $P\left(\mathcal{W}|D_{N}\right)$.
Therefore, PMFB-BackProp output tends to be more accurate - as averaging
the output of several MNNs is a common method to improve performance.
Additionally, the comparison between the performance of PMFB-BackProp
and MFB-BackProp usually allows us to estimate the level of certainty
in the estimate - since if the distribution becomes deterministic
($\left|h_{ij,l}^{\left(n\right)}\right|\rightarrow\infty$, so $P\left(W_{ij,l}=1|D_{n}\right)$
goes to $0$ or $1$) both must coincide. 

Lastly, to avoid a possible non-positive variance due to numerical
inaccuracy, we approximate (with a generally negligible error) $\sigma_{i\left(j\right),l}^{2}\approx\sigma_{i,l}^{2}+\mathrm{eps}$
where $\mathrm{eps}=2^{-52}$. Additionally, to avoid $G_{ij,L}\sim0\cdot\infty$
from generating nonsensical values ($\pm\infty$, NaN) when $\sigma_{i,L}\rightarrow0$,
we use instead the asymptotic form of $G_{ij,L}$ in that case: 
\begin{equation}
G_{ij,L}=-\frac{2\mu_{i\left(j\right),L}}{\sigma_{i,L}^{2}\sqrt{K_{L}}}\theta\left(-y_{i}\mu_{i\left(j\right),L}\right)\,.\label{eq: G asymptotic}
\end{equation}

\subsubsection*{Comparison with Backpropagation}

The increment in our algorithm (Eq.~\ref{eq: delta h}) is very similar
to the increment of the the weights in the BackPropagation algorithm
(BackProp) on a similar RMNN with converging architecture
\begin{equation}
\mathbf{v}_{L}=\mathrm{s}\left(\mathbf{W}_{L}\mathrm{s}\left(\mathbf{W}_{L-1}\mathrm{s}\left(\cdots\mathbf{W}_{1}\mathbf{v}_{0}\right)\right)\right),\label{eq: RMNN output}
\end{equation}
where $s\left(\cdot\right)$ is some sigmoid function. For example,
suppose a 2-layer BMNN \emph{i.e.}, $L=2$ (the following arguments
remain true for $L\geq2$). In this case, we have
\begin{align}
\Delta h_{ij,1} & =\frac{1}{2}\tanh\left[G_{i^{\prime}i,2}\right]y_{i^{\prime}}\left\langle W_{i^{\prime}i,2}\right\rangle \tanh\left[G_{ij,1}x_{j}\right]\label{eq: R 1}\\
\Delta h_{ij,2} & =\frac{1}{2}\tanh\left[G_{ij,2}\right]y_{i}\left\langle v_{j,1}\right\rangle \,.\label{eq:R 2}
\end{align}
For comparison, if we train a similar 2-layer converging RMNN (so
$\mathbf{v}_{2}=s\left(\mathbf{u}_{2}\right)$,$\mathbf{u}_{2}=\mathbf{W}_{2}\mathbf{v}_{1}$,
$\mathbf{v}_{1}=s\left(\mathbf{u}_{1}\right)$ and $\mathbf{u}_{1}=\mathbf{W}_{1}\mathbf{x}$)
with BackProp, the weights will update according to
\begin{eqnarray}
\Delta W_{ij,1} & = & -\eta_{1}\frac{\partial E}{\partial u_{i^{\prime},2}}W_{i^{\prime}i,2}s^{\prime}\left(u_{j,1}\right)x_{j}\label{eq: Backprop 1}\\
\Delta W_{ij,2} & = & -\eta_{2}\frac{\partial E}{\partial u_{i,2}}v_{j,1}\,,\label{eq: BackProp 2}
\end{eqnarray}
where $E$ is some non-negative error function and $\eta$ is a learning
rate. Notably, Eqs.~\ref{eq: R 1}-\ref{eq:R 2} can be written as
a somewhat modified version of Eqs.~\ref{eq: Backprop 1}-\ref{eq: BackProp 2},
with $E=-\sum_{i}\ln\Phi\left(y_{i}u_{i,L}/\left(\sqrt{K_{L}}\sigma_{i,L}\right)\right)$
and $s\left(u_{k,l}\right)=\left(2\Phi\left(u_{k,l}/\left(\sqrt{K_{l}}\sigma_{i,l}\right)\right)-1\right)$
(note $s\left(\cdot\right)$ has a sigmoid shape), and with $W_{ij,l}$,
$v_{i,l},$ and $u_{i,l}$ representing $\left\langle W_{ij,l}\right\rangle $,
$\left\langle v_{i,l}\right\rangle $ and $\sqrt{K_{l}}\mu_{i,l}$,
respectively. The modifications in Eqs.~\ref{eq: R 1}-\ref{eq:R 2}
in comparison to Eqs.~\ref{eq: Backprop 1}-\ref{eq: BackProp 2}
are: (1) the addition of the $\tanh\left(\cdot\right)$ functions
(including $\tanh\left(h_{ij,l}\right)=\left\langle W_{ij,l}\right\rangle $)
(2) $\mu_{i\left(j\right),l}$ replaces $\mu_{i,l}$ (3) $\sigma_{i,l}$
depends on the inputs and weights (second line on Eq.~\ref{eq: mu}).

Interestingly, this last property (3) entails that in the algorithm,
the input to each neuron is scaled adaptively. This implies that the
MFB-Backprop algorithm is invariant to changes in the amplitude of
of the input $\mathbf{x}$ (\emph{i.e.}, \textbf{$\mathbf{x}\rightarrow c\mathbf{x}$},
where $c>0$). This preserves the invariance of the output of the
BMNN to such amplitude changes in the input (the BMNN's invariance
can be seen from Eq.~\ref{eq: BMNN output}). Note that in standard
BackProp algorithm the performance is directly affected by the amplitude
of the input, so it is a recommended practice to re-scale it in pre-processing
\citep{LeCun2012}. In MFB-Backprop algorithm this becomes unnecessary.

Additionally, similarly to BackProp, if \textbf{$\mathbf{v}_{L}\rightarrow\mathbf{y}$
}then $E\rightarrow0$, and increments of the algorithm also go to
zero - since in this case $y_{i}\mu_{i\left(j\right),L}/\sigma_{i,l}\rightarrow\infty$,
and so $G_{ij,L}\rightarrow0$ (see Eqs.~\ref{eq: G} and \ref{eq: G asymptotic}).
This ``stable fixed point'' of the algorithm, corresponds to a deterministic
BMNN (in which$\left|h_{ij,l}^{\left(n\right)}\right|\rightarrow\infty$).
By ``stable fixed point'' we mean that near this point, any $\left\{ \mathbf{x}^{\left(n\right)},\mathbf{y}^{\left(n\right)}\right\} $
sample that does not contradict the estimated BMNN will not modify
this estimate (\emph{i.e.}, $\forall i,j,l:\,\Delta h_{ij,l}=0$),
since $G_{ij,L}=0$ if $y_{i}\mu_{i\left(j\right),L}>0$ in that case
(from Eq.~\ref{eq: G asymptotic}). Lastly, similarly to BackProp,
there are additional fixed points corresponding to highly symmetric
distributions. As a simple example, if $L\geq2$ and if $\forall i,j,l:\, h_{ij,l}^{\left(0\right)}=0$
(\emph{i.e.}, a uniform prior), than we will always have $\forall i,j,l:\,\Delta h_{ij,l}=0$.
Therefore, symmetry in the initial conditions can be deleterious.

\begin{algorithm}

\subsection*{Function $\mathcal{H}_{\mathrm{next}}=\mathrm{BayesUpdateStep}\left(\mathbf{x},\mathbf{y},\mathcal{H}\right)$}

\begin{algorithmic}

\STATE Initialize  $\nu_{k,0}=x_{k}$, $\Delta_{ij,L+1}=y_{j}$.

\STATE \% Forward-propagation phase

\FOR{ $m=1$ {\bfseries to} $L$}\STATE $\forall k$:
\begin{flalign*}
\mu_{k,m} & =\frac{1}{\sqrt{K_{m}}}\sum_{r\in K\left(k,m\right)}\tanh\left(h_{kr,m}\right)\nu_{r,m-1}\\
\sigma_{k,m}^{2} & =\frac{1}{K_{m}}\sum_{r\in K\left(k,m\right)}\left[\left(1-\nu_{r,m-1}^{2}\right)\left(1-\delta_{1m}\right)+\nu_{r,m-1}^{2}\mathrm{sech}^{2}\left(h_{kr,m}\right)+\mathrm{eps}\right]\\
\nu_{k,m} & =2\Phi\left(\mu_{k,m}/\sigma_{k,m}\right)-1
\end{flalign*}

\ENDFOR

\STATE \% Back-propagation phase

\FOR{ $l=L$ {\bfseries to} $1$} \STATE $\forall i,j$:
\begin{flalign*}
\mu_{i\left(j\right),l} & =\mu_{i,l}-\frac{1}{\sqrt{K_{l}}}\tanh\left(h_{ij,l}\right)\nu_{j,l-1}\\
G_{ij,l} & =\frac{2}{\sqrt{K_{l}}}\frac{\mathcal{N}\left(0|\mu_{i\left(j\right),l},\sigma_{i,l}^{2}\right)}{1+\left[\Phi\left(y_{i}\mu_{i\left(j\right),L}/\sigma_{i,L}\right)-1\right]\delta_{Ll}}
\end{flalign*}

\IF{$G_{ij,L}\notin\mathbb{R}$}

\STATE $G_{ij,L}=-2\frac{\mu_{i\left(j\right),L}}{\sigma_{i,L}^{2}\sqrt{K_{L}}}\theta\left(-y_{i}\mu_{i\left(j\right),L}\right)$

\ENDIF\STATE
\begin{eqnarray*}
\Delta_{ij,l} & = & \Delta_{i^{\prime}i,l+1}\tanh\left[G_{ij,l}\right]\tanh\left(h_{ij,l}\right)
\end{eqnarray*}

\STATE
\begin{eqnarray*}
R_{ij,l} & = & \Delta_{i^{\prime}i,l+1}\left[\delta_{l1}\tanh\left[G_{ij,l}x_{j}\right]+\left(1-\delta_{l1}\right)\tanh\left[G_{ij,l}\right]\nu_{j,l-1}\right]
\end{eqnarray*}

\ENDFOR

\STATE \% Update $\mathcal{H}$ 

\STATE $\forall l,k,r$:
\begin{flalign*}
h_{kr,l}^{\mathrm{next}} & =h_{kr,l}+\frac{1}{2}R_{kr,l}
\end{flalign*}
\end{algorithmic}\caption{A single update step of the the Mean Field Bayes Backpropagation (MFB-BackProp)
algorithm for BMNNs. The Matlab code for the algorithm will be available
at the author's website after publication. \label{alg:The-Variational-Bayes}}
\end{algorithm}

\section{Numerical Experiments}

\begin{figure*}
\begin{tabular}{ll}
\textbf{A} & \textbf{B}\tabularnewline
\includegraphics[width=0.47\textwidth]{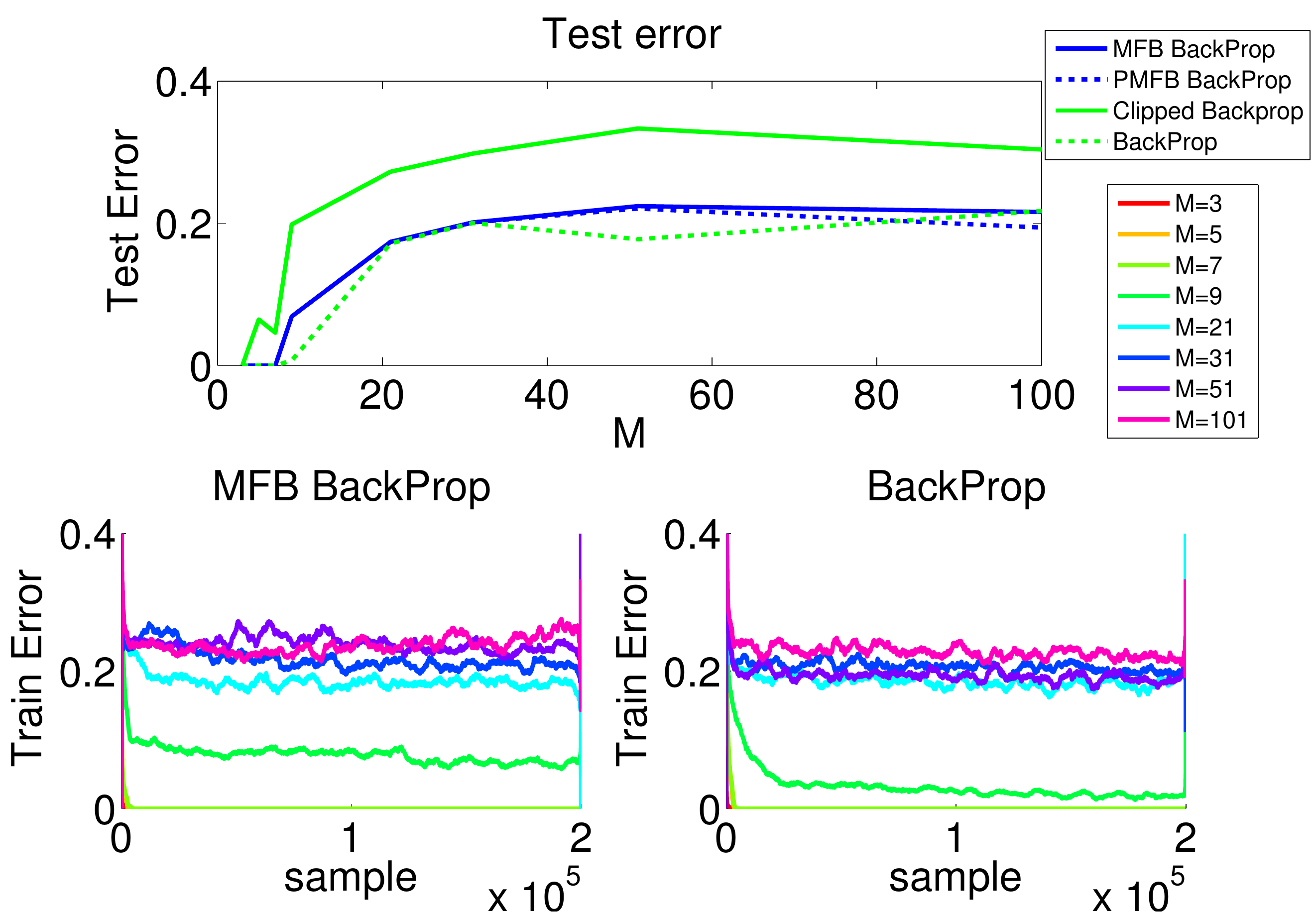} & \includegraphics[width=0.47\textwidth]{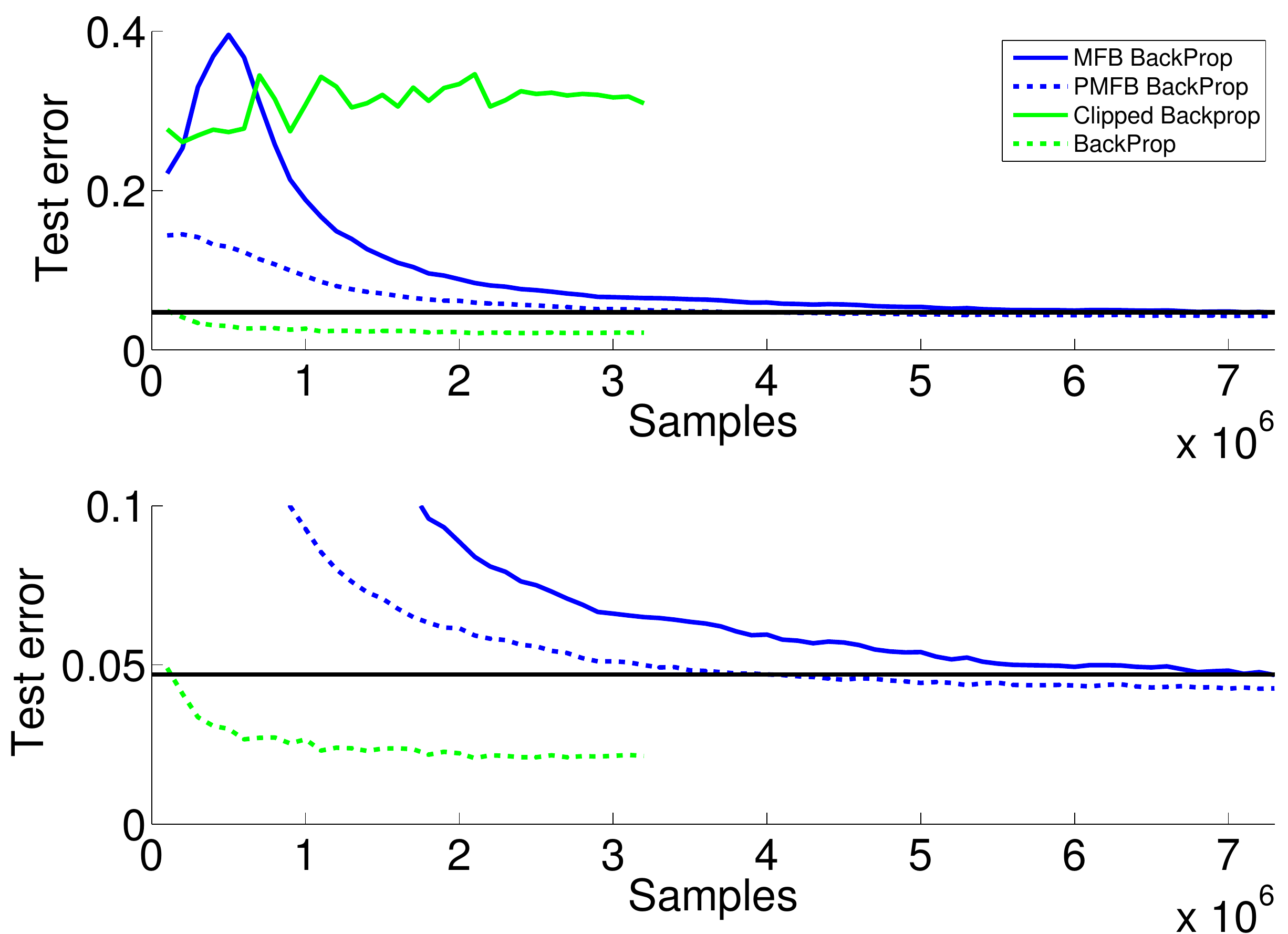}\tabularnewline
\end{tabular}

\caption{Performance of the MFB-BackProp (and PMFB-BackProp) algorithm on BMNN
in comparison with BackProp (and clipped BackProp) on a RMNN with
the same architecture. \textbf{(A) }A synthetic teacher student-problem
with random inputs. (\emph{bottom)} MFB-BackProp and BackProp show
error during training \emph{(top}) Mean test error on $10^{4}$ unseen
samples, after training has stopped\textbf{ (B) }Test error on the
MNIST data for a $785\times(301\times10)\times10$ 2-layer network.
For comparison we give the performance of BackProp on the same network,
as well as the performance (black line) of a fully connected ($785\times300\times10$)
RMNN from the literature \citep{Lecun1998}. The bottom graph ``zooms-in''
on the asymptotic behavior of the top figure for test error below
$0.1$. \label{fig:Performance-of-VBB} }
\end{figure*}

To test the MFB-BackProp algorithm, we run numerical experiments on
two tasks. The first is a simple teacher-student task with synthetic
data. The purpose of this teacher-student scenario is to test the
whether the algorithm works properly when (1) the labels are indeed
generated by the correct model class (a BMNN), as assumed by our Bayesian
framework (2) when the fan-in of the network is small, in contrast
to approximation 2. The second task is the standard MNIST handwritten
digits recognition task. In this task we examine the performance of
our algorithm for relatively large BMNN ($\left|\mathcal{W}\right|=2365860$,
while $N=60000$). The purpose of this task was to test whether the
algorithm can learn with real-valued inputs and in a more realistic
setting, in which the labels $\mathbf{y}$ do not arise from a BMNN
teacher, as assumed by the Bayesian approach. 

To facilitate comparison, we tested the several algorithms on each
task. The output of the network for each algorithms was: \textbf{(1)
MFB-BackProp} - the output of a BMNN (Eq.~\ref{eq: BMNN output})
with MAP weights (Eq.~\ref{eq: W MFB-BackProp}) estimated by MFB-BackProp
(Algorithm \ref{alg:The-Variational-Bayes}). \textbf{(2) PMFB-BackProp}
- the MAP output (Eq.~\ref{eq: y_p}) of a BMNNs estimated by MFB-BackProp
(Algorithm \ref{alg:The-Variational-Bayes}). \textbf{(3) BackProp}
- the output of an RMNN (Eq.~\ref{eq: RMNN output}), trained by
standard BackProp \citep{LeCun2012}. We use a RMNN with an identical
architecture to the BMNN we train with MFB-BackProp. The RMNN activation
function is $s\left(x\right)=1.7159\tanh\left(2x/3\right)$, as recommended
by \citep{LeCun2012} and we select the learning rate $\eta$ by a
parameter scan.\textbf{ (4) Clipped-BackProp} - the output of the
RMNN trained by BackProp, after we clip the weights to be binary (\emph{i.e.},$W_{ij,l}^{CBP}=\mathrm{sign}\left(W_{ij,l}^{BP}\right)$).

Note that only MFB-BackProp and Clipped-BackProp yield a MNN with
binary weights which can be used for hardware implementations. For
all algorithms we used uniform initial conditions, with std=1, as
recommended for BackProp \citep{LeCun2012} - $\sqrt{K_{l}/3}h_{ij,l}^{\left(0\right)}\sim\mathrm{U}\left[-1,1\right]$
for MFB-BackProp, and similarly for $W_{ij,l}^{\left(0\right)}$ in
BackProp.

\paragraph*{Synthetic Task Implementation.}

For all $M\in\left\{ 3,5,7,9,21,31,51,101\right\} $, we generated
synthetic data of $N=2\cdot10^{5}$ random binary samples $\mathbf{x}^{\left(n\right)}\in\left\{ -1,1\right\} ^{M}$
and labeled them with $\mathbf{y}^{\left(n\right)}\in\left\{ -1,1\right\} $
using a ``teacher'' - a 2-layer BMNN of size $M\times M\times1$.
We assumed the student knows the architecture of the teacher's BMNN.
However, the teacher's BMNN's weights are unknown, and were chosen
randomly before each trial of training. The student's task is to predict
the label for each $\mathbf{x}^{\left(n\right)}$, using the outputs
of the network for each algorithm. For each algorithm, the classification
error is $0$ if the output has the same sign as the label, and $1$
otherwise. During training, the classification error is averaged over
the last $5000$ samples. After training has stopped, the test error
is given by the mean error of the student on $N=10^{4}$ new random
samples of $\mathbf{x}^{\left(n\right)}$. For each $M$, we repeat
the training and testing for 10 trials. In Fig. \ref{fig:Performance-of-VBB}A
we present the trial with the best test error for each $M$. In the
top of Fig. \ref{fig:Performance-of-VBB}A we show the test error
of all algorithms we tested. The bottom panels of Fig. \ref{fig:Performance-of-VBB}A
show the performance of both the MFB-BackProp and BackProp during
training. The learning rate we used for Backprop (selected by a parameter
scan for each $M$) were: for $M=3$ $\eta=0.1$, for $M=5,7,9$ $\eta=3\cdot10^{-2}$,
for $M=21,31,51$ $\eta=3\cdot10^{-3}$ and for $M=101$ $\eta=10^{-3}$.

\paragraph*{Synthetic Task Results.}

For all algorithms the training error of the network output improves
during training (Fig. \ref{fig:Performance-of-VBB}A, \emph{bottom}).
Note that for small networks with $M\leq7$ the training error reaches
zero. As for the test error, MFB-BackProp, PMFB-BackProp and BackProp
all have a comparable performance, while Clipped-Backprop performs
significantly worse. These results indicate that MFB-BackProp can
work rather well, even when the assumption of a ``large fan-in''
is inaccurate (since $M$ was rather small in some cases).

\paragraph{MNIST Task Implementation.}

We tested the algorithms on the standard MNIST handwritten digits
database \citep{Lecun1998}. The training set contains $60,000$ images
($28\times28$ pixels) and the test set has other $10,000$ images.
The training set was presented repeatedly, each time with a randomized
order of samples. The task was to identify the $\mathrm{label}\in\left\{ 0,1,\dots9\right\} $,
using a BMNN classier trained by MFB-BackProp (as is standard, we
set $y_{k}=2\delta_{k,\mathrm{label}+1}-1$). In all networks we added
a constant 1 component to the input, to allow some (small) bias to
the neurons in the hidden layer ($V_{0}=785$). Also, we centralized
(removed the means) and normalized the input (so $\mathrm{std}=1$),
as recommended for BackProp \citep{LeCun2012}. Note that in this
task, we know that only a specific set of $10$ pattern is valid (out
of possible $2^{10}$). Therefore we need to decide how to classify
an image when the network output is ``illegal'' (\emph{e.g.}, $\left(-1,\dots,-1\right)$).
As standard for classification with RMNNs \citep{Lecun1998}, the
output neuron which has the highest input indicates the label of the
input pattern. We use this classification rule for all algorithms.
For BackProp we use $\eta=10^{-3}$, chosen by a parameter scan.

\paragraph{MNIST Task Results.}

A fully connected 2-layer ($785\times300\times10$) RMNN mentioned
in Lecun et al. \citeyearpar[Fig. 9]{Lecun1998}, trained using a
state-of-the-art \citep{LeCun2012} Levenberg-Marquardt algorithm,
achieved a $4.7\%$ test error. Our goal was to replicate this performance
using a BMNN with converging architecture, and so we used a wider
BMNN with $785\times(301\times10)\times10$ architecture. As can be
seen in Fig. \ref{fig:Performance-of-VBB}B, this goal was achieved,
since the final test errors were: MFB-BackProp - $4.68\%$, PMFB-BackProp
- $4.26\%$. Not surprisingly, the computational capability of a BMNN
is somewhat lower than a RMNN of a similar size, since the test error
for the RMNN with same architecture were: BackProp - $2.14\%$ and
Clipped BackProp - $30.95\%$. An increase in network width (or even
better, depth \citep{Siu1995}) is expected to further improve the
performance of the BMNN, but this is left to future work. Note the
performance of a SNN in the MNIST task is significantly worse when
the weights are constrained to be binary. For example, for a fully
connected SNN ($785\times10$), the test error is $29\%$ for binary
weights (using MFB-BackProp, not shown) and $12\%$ \citep{Lecun1998}
for real-valued weights. This demonstrates why our results for the
multilayer case are essential for obtaining good performance on realistic
tasks.

\section{Discussion}

Motivated by the recent success of MNNs, and the possibility of implementing
them in power-efficient hardware devices requiring limited parameter
precision, we developed a Bayesian algorithm for BMNNs. Specifically,
we derived the Mean Field-Bayes Backpropagation algorithm - a learning
algorithm for BMNNs in which we assumed a converging architecture
(\emph{i.e., }fan-out $1$, except for the input layer).

This online algorithm is essentially an analytic approximation to
the intractable Bayes calculation of the posterior distribution of
the weights after the arrival of a new data point. Note that this
is different from the common MNN training algorithms, such as BackProp,
which implement a minimization of some error function. To simplify
the intractable Bayes update rule we use two approximations. First,
we approximate the posterior using a product of its marginals (a `mean
field' approximation). Second, we assume the neuronal layers have
a ``large'' fan-in, so we can use the central limit theorem, as
well as first order approximations. After we obtain the approximated
updated posterior using the algorithm, it is trivial to find its maximum
(due to the posterior's factorized form).

To the best of our knowledge, this is the first training algorithm
for MNNs in general (\emph{i.e.}, not only binary) which is completely
Bayesian and scalable (\emph{i.e.}, one which does not use MCMC sampling
\citep{Mackay1992,Neal1995}). Despite its different origin, our algorithm
resembles the BackProp algorithm for RMNN - with a specific activation
function, learning rate and error function. However, there are a few
significant differences. For example, in our algorithm the input to
each neuron is scaled adaptively, preserving the amplitude invariance
of the BMNN. 

As far as we are aware, this is the first scalable algorithm for BMNNs.
Interestingly, in the special case of a single layer binary network,
our algorithm is almost identical to the online algorithms derived
in \citep{Sollaa,Ribeiro2011} (using similar Bayesian formalism and
approximations), and the ``greedy'' version of the belief-propagation
based algorithm derived in \citep{Braunstein}. The main difference
is the addition of the saturating function $\tanh\left(\cdot\right)$
in our algorithm. Had we used the $\mathrm{sign}\left(\cdot\right)$
function instead, we would have obtained the BPI algorithm \citep{Baldassi2007a}. 

We test numerically a BMNN trained using our algorithm in a synthetic
student-teacher task. The algorithm seems to work well even when the
network fan-in is small, in contrast to our assumptions. Next, we
test the algorithm on the standard MNIST handwritten digit classification
task. We demonstrate that the algorithm can train a 2-layer BMNN (with
$\sim10^{6}$ parameters), and achieve similar performance to 2-layer
RMNNs from the literature. In this task, the performance of the 2-layer
BMNN is comparable with BackProp tested on the same network (with
real weights), and significantly better than the clipped version of
BackProp (with binary weights). 

The numerical results suggest that 2-layer BMNNs can work just as
well as 2-layer RMNN, although they may require a larger width. The
weights of the BMNNs we have trained can now be immediately implemented
in a hardware chip, such as \citep{Karakiewicz2012}, significantly
improving their speed and energy efficiency in comparison to software-based
RMNNs. It remains to be seen whether deep BMNNs can compete with RMNNs
with (usually, fine tuned) deep architectures, which achieve state-of-the-art
performance. To do this, it would be desirable to lift the converging
architecture restriction. We hypothesize this constraint can be removed,
using the analogy of our algorithm with BackProp. We leave this for
future work, as well as quite a few other seemingly straightforward
extensions. For example, adapting our formalism to MNNs with discrete
weights \emph{- i.e.}, with more than 2 values. In the continuum limit,
such a generalization of the algorithm may be used for Bayesian training
of RMNNs.

\paragraph*{Acknowledgments}

The authors are grateful to C. Baldassi, A. Braunstein, and R. Zecchina
for helpful discussions and to T. Knafo for reviewing parts of this
manuscript. The research was partially funded by the Technion V.P.R.
fund and by the Intel Collaborative Research Institute for Computational
Intelligence (ICRI-CI).

\bibliographystyle{icml2014}
\bibliography{library}

\newpage{}

\appendix

\part*{Supplementary material - derivations \label{sec:Derivations}}

\section{The mean-field approximation\label{sec:The-mean-field-approximation}}

In this section we derive of Eqs.~\ref{eq: bayes update-1 - L} and
\ref{eq: P(y|x,W,D) - L}. Recall Eq.~\ref{eq: factorization assumption},
\[
\hat{P}\left(\mathcal{W}|D_{n}\right)=\prod_{i,j,l}\hat{P}\left(W_{ij,l}|D_{n}\right)\,,
\]
where $\hat{P}\left(\mathcal{W}|D_{n}\right)$ is an approximation
of $P\left(\mathcal{W}|D_{n}\right)$. In this section we answer the
following question - suppose we know $\hat{P}\left(\mathcal{W}|D_{n-1}\right)$.
How do we find $\hat{P}\left(\mathcal{W}|D_{n}\right)$? It is a standard
approximation to answer this question using a variational approach
(see \citet{Bishop2006}, and note that \citet{Sollaa,Ribeiro2011}
also used the same approach for a SNN with binary weights), through
the following two steps:
\begin{enumerate}
\item We use the Bayes update (Eq.~\ref{eq: bayes update}) with $\hat{P}\left(\mathcal{W}|D_{n-1}\right)$
as our prior 
\begin{eqnarray}
\tilde{P}\left(\mathcal{W}|D_{n}\right) & \propto & P\left(\mathbf{y}^{\left(n\right)}|\mathbf{x}^{\left(n\right)},\mathcal{W}\right)\hat{P}\left(\mathcal{W}|D_{n-1}\right)\nonumber \\
 & = & P\left(\mathbf{y}^{\left(n\right)}|\mathbf{x}^{\left(n\right)},\mathcal{W}\right)\prod_{i,j,l}\hat{P}\left(W_{ij,l}|D_{n-1}\right)\,,\label{eq: bayes Update}
\end{eqnarray}
where $\tilde{P}\left(\mathcal{W}|D_{n}\right)$ is some ``temporary''
posterior distribution.
\item We project $\hat{P}\left(\mathcal{W}|D_{n}\right)$ onto $\tilde{P}\left(\mathcal{W}|D_{n}\right)$
by minimizing the reverse Kullback-Leibler divergence (e.g., as in
the expectation propagation algorithm \citet{Minka2001,Bishop2006})
\[
D_{KL}\left(\tilde{P}\left(\mathcal{W}|D_{n}\right)||\hat{P}\left(\mathcal{W}|D_{n}\right)\right)=\sum_{\mathcal{W}}\tilde{P}\left(\mathcal{W}|D_{n}\right)\log\left(\frac{\tilde{P}\left(\mathcal{W}|D_{n}\right)}{\hat{P}\left(\mathcal{W}|D_{n}\right)}\right)
\]
with the normalization constraint $\sum_{W_{ij,l}}\hat{P}\left(W_{ij,l}|D_{n}\right)=1$
$\forall i,j,l$.
\end{enumerate}
The second step can be easily performed using Lagrange multipliers
\[
L\left(\hat{P}\left(\mathcal{W}|D_{n}\right)\right)=\sum_{\mathcal{W}^{\prime}}\tilde{P}\left(\mathcal{\mathcal{W}^{\prime}}|D_{n}\right)\log\left(\frac{\tilde{P}\left(\mathcal{W}^{\prime}|D_{n}\right)}{\prod_{k,r,m}\hat{P}\left(W_{kr,m}^{\prime}|D_{n}\right)}\right)+\sum_{k,r,m}\lambda_{kr,m}\left(1-\sum_{W_{kr,m}^{\prime}}\hat{P}\left(W_{kr,m}^{\prime}|D_{n}\right)\right)\,.
\]
The minimum is found by differentiating and equating to zero
\begin{eqnarray*}
0 & = & \frac{\partial L\left(\hat{P}\left(\mathcal{W}|D_{n}\right)\right)}{\partial\hat{P}\left(W_{ij,l}|D_{n}\right)}=-\frac{\sum_{\mathbf{\mathcal{W}}^{\prime}:W_{ij,l}^{\prime}=W_{ij,l}}\tilde{P}\left(\mathcal{W}^{\prime}|D_{n}\right)}{\hat{P}\left(W_{ij,l}|D_{n}\right)}-\lambda_{kr,m}\,.
\end{eqnarray*}
Using this equation together with the normalization constraint $\sum_{W_{kr,m}}\hat{P}\left(W_{kr,m}|D_{n}\right)=1$
$\forall k,r,m$ we obtain the result of the minimization through
marginalization 
\begin{eqnarray}
\hat{P}\left(W_{ij,l}|D_{n}\right) & = & \sum_{\mathbf{\mathcal{W}}^{\prime}:W_{ij,l}^{\prime}=W_{ij,l}}\tilde{P}\left(\mathcal{W}^{\prime}|D_{n}\right)\,,\label{eq: projection}
\end{eqnarray}
which is a known result \citep[p. 468]{Bishop2006}. Finally, we can
combine step 1 (Bayes update) with step 2 (projection) to a single
step
\begin{eqnarray*}
\hat{P}\left(W_{ij,l}|D_{n}\right) & = & \sum_{\mathbf{\mathcal{W}}^{\prime}:W_{ij,l}^{\prime}=W_{ij,l}}P\left(\mathbf{y}^{\left(n\right)}|\mathbf{x}^{\left(n\right)},\mathbf{\mathcal{W}}^{\prime}\right)\prod_{k,r,m}\hat{P}\left(W_{kr,m}^{\prime}|D_{n-1}\right)\,.
\end{eqnarray*}
This step is exactly Eqs.~\ref{eq: bayes update-1 - L} and \ref{eq: P(y|x,W,D) - L}
combined.

\section{Forward propagation of probabilities \label{sec:Forward-propagation-of}}

In this section we simplify the summations in Eqs.~(\ref{eq:P(v|v) - L}-\ref{eq:P(vm| W)}),
by assuming that the fan-in of all of the connections is ``large'',
\emph{i.e.}, $\forall m:K_{m}\rightarrow\infty$. Recall (from Eq.~\ref{eq:P(y|x,Wij,l) - L})
that $W_{ij,l}$ is a specific weight which is fixed (so $i,j$ and
$l$ are ``special'' indexes), while all the other weights $W_{kr,m}$
(for which $k\neq i$ or $r\neq j$ or $m\neq l$) are independent
binary random variables with $P\left(W_{kr,m}=1\right)=1-P\left(W_{kr,m}=-1\right),\,\forall k,r,m$.

\subsection{Some more preliminaries}

In addition to the notations defined in the the paper, we introduce
the following notation:
\begin{enumerate}
\item We use here a $x^{\pm}$ as a shorthand for $x=\pm1$ (so for example,
we write $P\left(y=1|x=-1\right)$ as $P\left(y^{+}|x^{-}\right)$. 
\item $\mathrm{Var}\left[x\right]=\left\langle X^{2}\right\rangle -\left\langle X\right\rangle ^{2}$
\end{enumerate}
Also, recall that for a binary variable $X$, we have 
\begin{eqnarray}
P\left(X=1\right) & = & P\left(X^{+}\right)=1-P\left(X^{-}\right)\nonumber \\
\left\langle X\right\rangle  & = & P\left(X^{+}\right)-P\left(X^{-}\right)=2P\left(X^{+}\right)-1\,.\label{eq: binary mean}\\
\left\langle X^{2}\right\rangle  & = & 1\nonumber \\
\mathrm{Var}\left[X\right] & = & 1-\left\langle X\right\rangle ^{2}\label{eq: binary variance}
\end{eqnarray}

\subsection{First layer}

In this subsection we calculate $P\left(\mathbf{v}_{1}\right)$, defined
in Eq.~\ref{eq: P(v1)}, assuming $l\neq1$. In this layer the input
is a fixed real vector $\mathbf{v}_{0}=\mathbf{x}\in\mathbb{R}^{V_{0}}$
(this is different from the next layers, where the input will be a
binary random vector).

Using Eqs.~\ref{eq: P(v1)} and \ref{eq:P(v|v) - L}, the fact that
the weights are binary and the central limit theorem we obtain 
\begin{eqnarray}
P\left(\mathbf{v}_{1}\right) & = & P\left(\mathbf{v}_{1}|\mathbf{v}_{0}=\mathbf{x}\right)\\
 & = & \sum_{\mathbf{W}_{1}^{\prime}}\prod_{k}\left[\theta\left(v_{k,1}\frac{1}{\sqrt{K_{1}}}\sum_{r\in K\left(k,1\right)}x_{r}W_{kr,1}^{\prime}\right)\prod_{r\in K\left(k,1\right)}P\left(W_{kr,1}^{\prime}\right)\right]\\
 & \approx & \prod_{k}\left[\int_{-\infty}^{\infty}\theta\left(v_{k,1}u_{k,1}\right)\mathcal{N}\left(u_{k,1}|\mu_{k,1},\sigma_{k,1}^{2}\right)du_{k,1}\right]\,,\label{eq:clt 1}\\
 & = & \prod_{k}\Phi\left(\frac{v_{k,1}\mu_{k,1}}{\sigma_{k,1}}\right)\,,\label{eq: prod Phi}
\end{eqnarray}
where in the approximated equality we changed the summation on $\mathbf{W}_{1}^{\prime}$
to a Gaussian integration on 
\[
u_{k,1}=\frac{1}{\sqrt{K_{1}}}\sum_{r\in K\left(k,1\right)}x_{r}W_{kr,1}^{\prime}\sim\mathcal{N}\left(\mu_{k,1},\sigma_{k,1}^{2}\right)\,,
\]
for which 
\begin{eqnarray}
\mu_{k,1} & = & \left\langle u_{k,m}\right\rangle \nonumber \\
 & = & \left\langle \frac{1}{\sqrt{K_{1}}}\sum_{r\in K\left(k,1\right)}x_{r}W_{kr,1}^{\prime}\right\rangle \\
 & = & \frac{1}{\sqrt{K_{1}}}\sum_{r\in K\left(k,1\right)}\left\langle W_{kr,1}\right\rangle x_{k}\,,\label{eq: mu for l=00003D1}\\
\sigma_{k,1}^{2} & = & \mathrm{Var}\left[u_{k,m}\right]\nonumber \\
 & = & \mathrm{Var}\left[\frac{1}{\sqrt{K_{1}}}\sum_{r\in K\left(k,1\right)}x_{r}W_{kr,1}^{\prime}\right]\\
 & = & \frac{1}{K_{1}}\sum_{r\in K\left(k,1\right)}x_{r}^{2}\mathrm{Var}\left[W_{kr,1}^{\prime}\right]\\
 & = & \frac{1}{K_{1}}\sum_{r\in K\left(k,1\right)}x_{r}^{2}\left(1-\left\langle W_{kr,1}\right\rangle ^{2}\right),\label{eq: sigma for l=00003D1}
\end{eqnarray}
where we used Eq.~\ref{eq: binary variance} (recall also that $W^{\pm}$
means $W=\pm1$). Note that from Eq.~\ref{eq: prod Phi} we obtained
that the outputs of the first layer are independent 
\begin{equation}
P\left(\mathbf{v}_{1}\right)=\prod_{k}P\left(v_{k,1}\right)=\prod_{k}\Phi\left(v_{k,1}\mu_{k,1}/\sigma_{k,1}\right)\,.\label{eq: P(v1) independent}
\end{equation}
Also note, that if $\left|\mu_{k,1}\right|\gg\sigma_{k,1}$, our CLT-based
approximation in Eq.~\ref{eq:clt 1} breaks down. Despite this, in
this limit, $P\left(v_{k,1}\right)\approx\theta\left(v_{k,1}\mu_{k,1}\right)$
with or without the approximation. However, is is important to note
that the asymptotic form at which we approach this limit is different
without the CLT-based approximation.

\subsection{Layers $m=2,3,\dots,l$}

In this subsection we calculate $P\left(\mathbf{v}_{m}\right)$ $\forall m\in\left\{ 2,..,l-1\right\} $,
as defined in Eq.~\ref{eq:P(v|x)}. In this layer the input is a
random binary vector $\mathbf{v}_{m-1}\in\left\{ -1,1\right\} ^{V_{m-1}}$.
From Eq.~\ref{eq: P(v1) independent} we know that the inputs are
independent for $m=2$ (\emph{i.e.}, $P\left(\mathbf{v}_{1}\right)=\prod_{k}P\left(v_{k,1}\right)$).
We assume this is true $\forall m$, and this proved by induction
($i.e.$, assuming $P\left(\mathbf{v}_{m}\right)=\prod_{k}P\left(v_{k,m}\right)$
will yield $P\left(\mathbf{v}_{m+1}\right)=\prod_{k}P\left(v_{k,m+1}\right)$).

Using Eqs.~\ref{eq:P(v|v) - L} and \ref{eq:P(v|x)}, we can perform
very similar calculation for as we did for the first layer
\begin{eqnarray}
P\left(\mathbf{v}_{m}\right) & = & \sum_{\mathbf{v}_{m-1}}\sum_{\mathbf{W}_{m}^{\prime}}\prod_{k}\left[\theta\left(v_{k,m}\frac{1}{\sqrt{K_{m}}}\sum_{r\in K\left(k,m\right)}v_{m-1}W_{kr,m}^{\prime}\right)\prod_{r\in K\left(k,m\right)}P\left(W_{kr,m}^{\prime}\right)P\left(v_{r,m-1}\right)\right]\nonumber \\
 & \approx & \prod_{k}\left[\int_{-\infty}^{\infty}\theta\left(v_{k,m}u_{k,m}\right)\mathcal{N}\left(u_{k,m}|\mu_{k,m},\sigma_{k,m}^{2}\right)du_{k,m}\right]\,,\label{eq:clt 2}\\
 & = & \prod_{k}\Phi\left(\frac{v_{k,m}\mu_{k,m}}{\sigma_{k,m}}\right)\,,\nonumber 
\end{eqnarray}
where we similarly approximated 
\[
u_{k,m}=\frac{1}{\sqrt{K_{m}}}\sum_{r\in K\left(k,m\right)}v_{r,m-1}W_{kr,m}^{\prime}\sim\mathcal{N}\left(\mu_{k,m},\sigma_{k,m}^{2}\right)\,,
\]

for which

\begin{eqnarray}
\mu_{k,m} & = & \left\langle \frac{1}{\sqrt{K_{m}}}\sum_{r\in K\left(k,m\right)}v_{r,m-1}W_{kr,m}^{\prime}\right\rangle \nonumber \\
 & = & \frac{1}{\sqrt{K_{m}}}\sum_{r\in K\left(k,m\right)}\left\langle W_{kr,m}\right\rangle \left\langle v_{r,m-1}\right\rangle \,,\label{eq: mu-1}\\
\sigma_{k,m}^{2} & = & \mathrm{Var}\left[\frac{1}{\sqrt{K_{m}}}\sum_{r\in K\left(k,m\right)}v_{r,m-1}W_{kr,m}^{\prime}\right]\nonumber \\
 & = & \frac{1}{K_{m}}\sum_{r\in K\left(k,m\right)}\mathrm{Var}\left[v_{r,m-1}W_{kr,m}^{\prime}\right]\nonumber \\
 & = & \frac{1}{K_{m}}\sum_{r\in K\left(k,m\right)}\left(1-\left\langle v_{r,m-1}\right\rangle ^{2}\left\langle W_{kr,m}\right\rangle ^{2}\right)\,,\label{eq:sigma}
\end{eqnarray}
with 
\begin{equation}
\left\langle v_{r,m-1}\right\rangle =2P\left(v_{k,m-1}^{+}\right)-1=2\Phi\left(\mu_{k,m}/\sigma_{k,m}\right)-1\,.\label{eq: <v | x  >}
\end{equation}
Again, if $\left|\mu_{k,m}\right|\gg\sigma_{k,m}$, our CLT-based
approximation in Eq.~\ref{eq:clt 2} breaks down, but $P\left(v_{k,m}\right)\approx\theta\left(v_{k,m}\mu_{k,m}\right)$
holds even then. Again, the asymptotic form at which we approach this
limit is different without the CLT-based approximation. Also, we got
that the outputs are independent $P\left(\mathbf{v}_{m}\right)=\prod_{k}P\left(v_{k,m}\right)$,
so we can similarly calculate $P\left(\mathbf{v}_{m+1}\right)$ for
$\forall m>1$.

\subsection{Layer $l$}

In this subsection we calculate $P\left(\mathbf{v}_{l}|W_{ij,l}\right)$,
as defined in Eq.~\ref{eq:P(vl | W)}. We will have to consider two
different cases here - when $l=1$ and when $l\neq1$, since we have
different type of inputs to the layer in each case. Note that since
$W_{ij,l}$ is a ``special'' fixed weight (see Eq.~\ref{eq:P(y|x,Wij,l) - L}),
we will have to redefine $\mathbf{u}_{l}$ (originally defined in
Eq.~\ref{eq: CLT}) as if $W_{ij,l}$ ``disconnected''.

\subsubsection{Case when $l\neq1$}

First, we re-define \textbf{$\mathbf{u}_{l}$} in this layer, so that
\[
u_{k,l}=\begin{cases}
\frac{1}{\sqrt{K_{l}}}\sum_{r\in K\left(k,l\right)}v_{r,l-1}W_{kr,l}^{\prime}\,\,\,\sim\mathcal{N}\left(\mu_{k,l},\sigma_{k,l}^{2}\right) & ,\,\mathrm{if}\,\, k\neq i\\
\frac{1}{\sqrt{K_{l}}}\sum_{r\in K\left(k,l\right)\backslash j}v_{r,l-1}W_{kr,l}^{\prime}\,\sim\mathcal{N}\left(\mu_{i\left(j\right),l},\sigma_{i\left(j\right),l}^{2}\right) & ,\,\mathrm{if}\,\, k=i
\end{cases}\,.
\]
Now, $\mu_{k,l}$ and $\sigma_{k,l}$ are defined as in Eqs.~\ref{eq: mu-1}-\ref{eq:sigma},
while 
\begin{eqnarray}
\mu_{i\left(j\right),l} & = & \left\langle u_{i,l}\right\rangle \nonumber \\
 & = & \frac{1}{\sqrt{K_{l}}}\sum_{r\in K\left(i,l\right)\backslash j}\left\langle W_{ir,l}\right\rangle \left\langle v_{r,l-1}\right\rangle \,,\label{eq: mu 2}\\
\sigma_{i\left(j\right),l}^{2} & = & \mathrm{Var}\left[u_{i,l}\right]\nonumber \\
 & = & \frac{1}{K_{l}}\sum_{r\in K\left(i,l\right)\backslash j}\left(1-\left\langle v_{r,l-1}\right\rangle ^{2}\left\langle W_{ir,l}\right\rangle ^{2}\right)\,,\label{eq: sigma 2}
\end{eqnarray}
Next, using similar methods (as we did before) on Eq.~\ref{eq:P(vl | W)}
(with $P\left(\mathbf{v}_{l}|\mathbf{v}_{l-1},W_{ij,l}\right)$ defined
immediately after Eq.~\ref{eq:P(v|v) - L}), we have%
\footnote{In the next equation we say that $P\left(W_{ij,l}\right)/P\left(W_{ij,l}\right)=1$
even if $P\left(W_{ij,l}\right)=0$, so that $\left(P\left(W_{ij,l}\right)\right)^{-1}\prod_{r\in K\left(i,l\right)}P\left(W_{ir,l}^{\prime}\right)=\prod_{r\in K\left(i,l\right)\backslash j}P\left(W_{ir,l}^{\prime}\right)$ %
} 

\begin{eqnarray}
 &  & P\left(\mathbf{v}_{l}|W_{ij,l}\right)\nonumber \\
 & = & \sum_{\mathbf{v}_{l-1}}\sum_{\mathbf{W}_{l}^{\prime}:W_{ij,l}^{\prime}=W_{ij,l}}\left[\left(P\left(W_{ij,l}\right)\right)^{-1}\prod_{k}\theta\left(v_{k,l}\frac{1}{\sqrt{K_{l}}}\sum_{r\in K\left(k,l\right)}v_{r,l-1}W_{kr,l}^{\prime}\right)\prod_{r\in K\left(k,l\right)}P\left(W_{kr,l}^{\prime}\right)P\left(v_{r,l-1}\right)\right]P\left(v_{j,l-1}\right)\nonumber \\
 & \approx & \left[\prod_{k\neq i}\Phi\left(\frac{v_{k,l}\mu_{k,l}}{\sigma_{k,l}}\right)\right]\sum_{v_{j,l-1}}\Phi\left(\frac{v_{i,l}}{\sigma_{i\left(j\right),l}}\left(\mu_{i\left(j\right),l}+\frac{1}{\sqrt{K_{l}}}W_{ij,l}v_{j,l-1}\right)\right)P\left(v_{j,l-1}\right)\,\label{eq: clt 3}\\
 & = & \left[\prod_{k\neq i}P\left(v_{k,l}\right)\right]P\left(v_{i,l}|W_{ij,l}\right)\,.
\end{eqnarray}
If $\sigma_{i\left(j\right),l}\gg1/\sqrt{K_{l}}$ (which can be reasonable
if $K_{l}$ is large) then we can perform a first order Taylor expansion
of the last line and obtain
\begin{eqnarray}
P\left(v_{i,l}|W_{ij,l}\right) & = & \sum_{v_{j,l-1}}\Phi\left(\frac{v_{i,l}}{\sigma_{i\left(j\right),l}}\left(\mu_{i\left(j\right),l}+\frac{1}{\sqrt{K_{l}}}W_{ij,l}v_{j,l-1}\right)\right)P\left(v_{j,l-1}\right)\nonumber \\
 & \approx & \sum_{v_{j,l-1}}\left[\Phi\left(\frac{v_{i,l}\mu_{i\left(j\right),l}}{\sigma_{i\left(j\right),l}}\right)+\frac{1}{\sqrt{K_{l}}}\mathcal{N}\left(0|\mu_{i\left(j\right),l},\sigma_{i\left(j\right),l}^{2}\right)W_{ij,l}v_{i,l}v_{j,l-1}\right]P\left(v_{j,l-1}\right)\\
 & = & \Phi\left(\frac{v_{i,l}\mu_{i\left(j\right),l}}{\sigma_{i\left(j\right),l}}\right)+\frac{1}{\sqrt{K_{l}}}\mathcal{N}\left(0|\mu_{i\left(j\right),l},\sigma_{i\left(j\right),l}^{2}\right)W_{ij,l}v_{i,l}\left\langle v_{j,l-1}\right\rangle \,.\label{eq: case 1}
\end{eqnarray}
However, if $\left|\mu_{i\left(j\right),l}\right|\ll1/\sqrt{K_{l}}$
\emph{and} $\sigma_{i\left(j\right),l}\ll1/\sqrt{K_{l}}$ we can obtain
instead (even though the CLT is not valid)

\begin{eqnarray}
P\left(v_{i,l}|W_{ij,l}\right) & = & \sum_{v_{j,l-1}}\theta\left(v_{i,l}W_{ij,l}v_{j,l-1}\right)P\left(v_{j,l-1}\right)\label{eq: case 2}\\
 & = & \frac{1}{2}\left(1+W_{ij,l}v_{i,l}\left\langle v_{j,l-1}\right\rangle \right)\,.
\end{eqnarray}
Note that this separation of the last two limit cases is particularly
important, since if we use Eq.~\ref{eq: case 1} in the second case,
the equation will diverge. Both limit cases can be heuristically
combined into one equation

\begin{eqnarray}
P\left(v_{i,l}|W_{ij,l}\right) & \approx & \Phi\left(\frac{v_{i,l}\mu_{i\left(j\right),l}}{\sigma_{i\left(j\right),l}}\right)+\frac{1}{2}\tanh\left[\frac{2}{\sqrt{K_{l}}}\mathcal{N}\left(0|\mu_{i\left(j\right),l},\sigma_{i\left(j\right),l}^{2}\right)\right]W_{ij,l}v_{i,l}\left\langle v_{j,l-1}\right\rangle \,,\label{eq: P(v_l | x W)}
\end{eqnarray}
since

\[
\tanh\left(x\right)\approx\begin{cases}
x & ,\,\,\mathrm{if}\,\left|x\right|\ll1\\
\mathrm{sign\left(x\right)} & ,\,\,\mathrm{if}\,\left|x\right|\gg1
\end{cases}\,.
\]
Importantly, since the two limit cases were ``glued'' heuristically,
Eq.~\ref{eq: P(v_l | x W)} does not give the asymptotic form at
which we approach the limit $\left|\mu_{i\left(j\right),l}\right|\ll1/\sqrt{K_{l}}$
\emph{and} $\sigma_{i\left(j\right),l}\ll1/\sqrt{K_{l}}$. Also, if
$\left|\mu_{i\left(j\right),l}\right|\gg\sigma_{i\left(j\right),l}$
and also $\left|\mu_{i\left(j\right),l}\right|>1/\sqrt{K_{l}}$ the
assumptions behind Eq.~\ref{eq: P(v_l | x W)} break down, but the
result $P\left(v_{i,l}|W_{ij,l}\right)\approx\theta\left(v_{i,l}\mu_{i\left(j\right),l}\right)$
remains. Again, the asymptotic form at which we approach this limit
is different without the CLT-based approximation.

\subsubsection{Case when $l=1$}

In this case, we re-define \textbf{$\mathbf{u}_{l}$} in this layer,
so that 
\[
u_{k,l}=\begin{cases}
\frac{1}{\sqrt{K_{l}}}\sum_{r\in K\left(k,l\right)}x_{r,}W_{kr,l}^{\prime}\,\,\,\sim\mathcal{N}\left(\mu_{k,l},\sigma_{k,l}^{2}\right) & ,\,\mathrm{if}\,\, k\neq i\\
\frac{1}{\sqrt{K_{l}}}\sum_{r\in K\left(k,l\right)\backslash j}x_{r}W_{kr,l}^{\prime}\,\sim\mathcal{N}\left(\mu_{i\left(j\right),l},\sigma_{i\left(j\right),l}^{2}\right) & ,\,\mathrm{if}\,\, k=i
\end{cases}
\]
with $\mu_{k,l}$ and $\sigma_{k,l}$ are defined as in Eqs.~\ref{eq: mu for l=00003D1}-\ref{eq: sigma for l=00003D1},
while 
\begin{eqnarray}
\mu_{i\left(j\right),l} & = & \frac{1}{\sqrt{K_{l}}}\sum_{r\in K\left(i,l\right)\backslash j}\left\langle W_{ir,l}\right\rangle x_{r}\,,\label{eq: mu 2-1}\\
\sigma_{i\left(j\right),l}^{2} & = & \frac{1}{K_{l}}\sum_{r\in K\left(i,l\right)\backslash j}x_{r}^{2}\left(1-\left\langle W_{ir,l}\right\rangle ^{2}\right)\,.\label{eq: sigma 2-1}
\end{eqnarray}
Thus, we obtain
\begin{eqnarray*}
 &  & P\left(\mathbf{v}_{l}|W_{ij,l}\right)\\
 & = & \sum_{\mathbf{W}_{l}^{\prime}:W_{ij,l}^{\prime}=W_{ij,l}}\left[\left(P\left(W_{ij,l}\right)\right)^{-1}\prod_{k}\theta\left(v_{k,l}\frac{1}{\sqrt{K_{l}}}\sum_{r\in K\left(k,l\right)}x_{r,l-1}W_{kr,l}^{\prime}\right)\prod_{r\in K\left(k,l\right)}P\left(W_{kr,l}^{\prime}\right)\right]\\
 & \approx & \left[\prod_{k\neq i}\Phi\left(\frac{v_{k,l}\mu_{k,l}}{\sigma_{k,l}}\right)\right]\Phi\left(\frac{v_{i,l}}{\sigma_{i\left(j\right),l}}\left(\mu_{i\left(j\right),l}+\frac{1}{\sqrt{K_{l}}}W_{ij,l}x_{j}\right)\right)\\
 & = & \left[\prod_{k\neq i}P\left(v_{k,l}\right)\right]P\left(v_{i,l}|W_{ij,l}\right)\,.
\end{eqnarray*}
If $\left|\mu_{i\left(j\right),l}\right|\gg x_{j}/\sqrt{K_{l}}$ \emph{or}
$\sigma_{i\left(j\right),l}\gg x_{j}/\sqrt{K_{l}}$ then we can perform
a first order Taylor expansion of the last line and obtain
\begin{equation}
P\left(v_{i,l}|W_{ij,l}\right)=\Phi\left(\frac{v_{i,l}\mu_{i\left(j\right),l}}{\sigma_{i\left(j\right),l}}\right)+\frac{1}{\sqrt{K_{l}}}\mathcal{N}\left(0|\mu_{i\left(j\right),l},\sigma_{i\left(j\right),l}^{2}\right)W_{ij,l}v_{i,l}x_{j}\,.\label{eq: case 1-1}
\end{equation}
and if $\left|\mu_{i\left(j\right),l}\right|\ll x_{j}/\sqrt{K_{l}}$
\emph{and} $\sigma_{i\left(j\right),l}\ll x_{j}/\sqrt{K_{l}}$, we
obtain instead

\[
P\left(v_{i,l}|W_{ij,l}\right)=\frac{1}{2}\left(1+v_{i,l}W_{ij,l}\mathrm{sign}\left(x_{j}\right)\right)
\]
Both limit cases can be again combined into a single equation

\begin{eqnarray}
P\left(v_{i,l}|W_{ij,l}\right) & \approx & \Phi\left(\frac{v_{i,l}\mu_{i\left(j\right),l}}{\sigma_{i\left(j\right),l}}\right)+\frac{1}{2}\tanh\left[\frac{2}{\sqrt{K_{l}}}\mathcal{N}\left(0|\mu_{i\left(j\right),l},\sigma_{i\left(j\right),l}^{2}\right)x_{j}\right]W_{ij,l}v_{i,l}\,.\label{eq: P(v_l | x W)-1}
\end{eqnarray}
Again, if $\left|\mu_{i\left(j\right),l}\right|\gg\sigma_{i\left(j\right),l}$,
the assumptions behind Eq.~\ref{eq: P(v_l | x W)} breaks down, but
in that limit we still have $P\left(v_{i,l}|W_{ij,l}\right)\approx\theta\left(v_{i,l}\left(\mu_{i\left(j\right),l}+W_{ij,l}x_{j}/\sqrt{K_{l}}\right)\right)$.
Again, the asymptotic form at which we approach this limit is different
without the CLT-based approximation.

\subsection{Layers $m=l+1,l+2\dots,L$}

In this subsection we calculate $P\left(\mathbf{v}_{m}|W_{ij,l}\right),$
$\forall m\geq l$, as defined in Eq.~\ref{eq:P(vm| W)}. We denote
$c\left(i,l,l^{\prime}\right)$ ('child') to be the index of the neuron
in the $l^{\prime}$ layer which is receiving input (through other
neurons) from the $i$-th neuron in the $l$ layer, and $i_{m}=c\left(i,l,m\right)$
and $j_{m}=c\left(i,l,m-1\right)$. Using similar methods (as before)
on Eqs.~\ref{eq:P(v|v) - L} and \ref{eq:P(vm| W)}, we obtain
\[
P\left(\mathbf{v}_{m}|W_{ij,l}\right)=\left[\prod_{k\neq i_{m}}\Phi\left(\frac{v_{k,m}\mu_{k,m}}{\sigma_{k,m}}\right)\right]P\left(v_{i_{m},m}|W_{ij,l}\right)
\]
with
\begin{eqnarray}
 &  & P\left(v_{i_{m},m}|W_{ij,l}\right)\nonumber \\
 & = & \sum_{\mathbf{v}_{m-1},\mathbf{W}_{m}^{\prime}}\theta\left(v_{i_{m},m}\frac{1}{\sqrt{K_{m}}}\sum_{r\in K\left(i_{m},m\right)}v_{r,m-1}W_{i_{m}r,m}^{\prime}\right)\prod_{r\in K\left(i_{m},m\right)}P\left(W_{i_{m}r,m}^{\prime}\right)\prod_{r\in K\left(i_{m},m\right)\backslash j_{m}}P\left(v_{r,m-1}\right)P\left(v_{j_{m},m-1}|W_{ij,l}\right)\nonumber \\
 & \approx & \sum_{v_{j_{m},m-1},W_{i_{m}j_{m},m}^{\prime}}\Phi\left(\frac{v_{i_{m},m}}{\sigma_{i_{m}\left(j_{m}\right),m}}\left(\mu_{i_{m}\left(j_{m}\right),m}+\frac{1}{\sqrt{K_{m}}}W_{i_{m}j_{m},m}v_{j_{m},m-1}\right)\right)P\left(W_{i_{m}j_{m},m}^{\prime}\right)P\left(v_{j_{m},m-1}|W_{ij,l}\right)\,,\label{eq:clt 4}\\
 & \approx & \Phi\left(\frac{v_{i_{m},m}\mu_{i_{m}\left(j_{m}\right),m}}{\sigma_{i_{m}\left(j_{m}\right),m}}\right)+\frac{1}{2}\tanh\left[\frac{2}{\sqrt{K_{m}}}\mathcal{N}\left(0|\mu_{i_{m}\left(j_{m}\right),m},\sigma_{i_{m}\left(j_{m}\right),m}^{2}\right)\right]\left\langle W_{i_{m}j_{m},m}\right\rangle v_{i_{m},m}\left\langle v_{j_{m},m-1}|W_{ij,l}\right\rangle \label{eq:P(v_m | x , W)}
\end{eqnarray}
where $\mu_{k,m}$ and $\sigma_{k,m}$ are defined as in Eqs.~\ref{eq: mu-1}-\ref{eq:sigma},
$\mu_{i_{m}\left(j_{m}\right),m}$ and $\sigma_{i_{m}\left(j_{m}\right),m}$
are defined as in Eqs.~\ref{eq: mu 2}-\ref{eq: sigma 2}  and again
we divided into two limits, as in Eq.~\ref{eq: P(v_l | x W)}. Also,
if $\left|\mu_{i_{m}\left(j_{m}\right),m}\right|\gg\sigma_{i_{m}\left(j_{m}\right),m}$
and $\left|\mu_{i_{m}\left(j_{m}\right),m}\right|>1/\sqrt{K_{m}}$
our assumptions break down but we still have $P\left(v_{i_{m},m}|W_{ij,l}\right)\approx\theta\left(v_{i_{m},m}\mu_{i_{m}\left(j_{m}\right),m}\right)$.
Again, the asymptotic form at which we approach this limit is different
without the CLT-based approximation.

\section{The log-likelihood ratio\label{sec:The-log-likelihood-ratio}}

Using the results from the previous section, we can now calculate
$P\left(\mathbf{y}|W_{ij,l}\right)=P\left(\mathbf{v}_{L}=\mathbf{y}|W_{ij,l}\right)$
for every $W_{ij,l}$ . We instead calculate the following, equivalently
useful, log-likelihood ratio
\begin{equation}
R_{ij,l}=\ln\frac{P\left(\mathbf{y}|W_{ij,l}^{+}\right)}{P\left(\mathbf{y}|W_{ij,l}^{-}\right)}\,.\label{eq: R_ij,l}
\end{equation}
This quantity is useful, since, from Eq.~\ref{eq: bayes update-1 - L},
it uniquely determines the Bayes updates of the posterior as 
\[
\ln\frac{P\left(W_{ij,l}^{+}|D_{n}\right)}{P\left(W_{ij,l}^{-}|D_{n}\right)}=\ln\frac{P\left(W_{ij,l}^{+}|D_{n-1}\right)}{P\left(W_{ij,l}^{-}|D_{n-1}\right)}+R_{ij,l}^{\left(n\right)}\,.
\]
Next, we assume that $R_{ij,l}\ll1$ and discuss the complementary
case towards the end of this section. If $l=L=1$, then, using Eq.~\ref{eq: P(v_l | x W)-1},
we find
\begin{eqnarray}
R_{ij,l} & = & \ln\frac{\Phi\left(\frac{y_{i}\mu_{i\left(j\right),L}}{\sigma_{i\left(j\right),L}}\right)+\frac{1}{2}\tanh\left[\frac{2}{\sqrt{K_{L}}}\mathcal{N}\left(0|\mu_{i\left(j\right),L},\sigma_{i\left(j\right),L}^{2}\right)x_{j}y_{i}\right]}{\Phi\left(\frac{y_{i}\mu_{i\left(j\right),L}}{\sigma_{i\left(j\right),L}}\right)-\frac{1}{2}\tanh\left[\frac{2}{\sqrt{K_{L}}}\mathcal{N}\left(0|\mu_{i\left(j\right),L},\sigma_{i\left(j\right),L}^{2}\right)x_{j}y_{i}\right]}\nonumber \\
 & \approx & \ln\frac{1+\frac{1}{2}\tanh\left[\frac{2\mathcal{N}\left(0|\mu_{i\left(j\right),L},\sigma_{i\left(j\right),L}^{2}\right)}{\sqrt{K_{L}}\Phi\left(y_{i}\mu_{i\left(j\right),L}/\sigma_{i\left(j\right),L}\right)}x_{j}y_{i}\right]}{1-\frac{1}{2}\tanh\left[\frac{2\mathcal{N}\left(0|\mu_{i\left(j\right),L},\sigma_{i\left(j\right),L}^{2}\right)}{\sqrt{K_{L}}\Phi\left(y_{i}\mu_{i\left(j\right),L}/\sigma_{i\left(j\right),L}\right)}x_{j}y_{i}\right]}\\
 & \approx & \tanh\left[\frac{2}{\sqrt{K_{L}}}\frac{\mathcal{N}\left(0|\mu_{i\left(j\right),L},\sigma_{i\left(j\right),L}^{2}\right)}{\Phi\left(y_{i}\mu_{i\left(j\right),L}/\sigma_{i\left(j\right),L}\right)}x_{j}y_{i}\right]\,.\label{eq: Delta L-l-1}
\end{eqnarray}
where we used a first order Taylor expansion, assuming $R_{ij,l}\ll1$.

If $l=L>1$, then, using Eq.~\ref{eq: P(v_l | x W)}, we find

\begin{eqnarray}
R_{ij,l} & = & \ln\frac{\Phi\left(\frac{y_{i}\mu_{i\left(j\right),L}}{\sigma_{i\left(j\right),L}}\right)+\frac{1}{2}\tanh\left[\frac{2}{\sqrt{K_{L}}}\mathcal{N}\left(0|\mu_{i\left(j\right),L},\sigma_{i\left(j\right),L}^{2}\right)y_{i}\right]\left\langle v_{j,L-1}\right\rangle }{\Phi\left(\frac{y_{i}\mu_{i\left(j\right),L}}{\sigma_{i\left(j\right),L}}\right)-\frac{1}{2}\tanh\left[\frac{2}{\sqrt{K_{L}}}\mathcal{N}\left(0|\mu_{i\left(j\right),L},\sigma_{i\left(j\right),L}^{2}\right)y_{i}\right]\left\langle v_{j,L-1}\right\rangle }\nonumber \\
 & \approx & \tanh\left[\frac{2}{\sqrt{K_{L}}}\frac{\mathcal{N}\left(0|\mu_{i\left(j\right),L},\sigma_{i\left(j\right),L}^{2}\right)}{\Phi\left(y_{i}\mu_{i\left(j\right),L}/\sigma_{i\left(j\right),L}\right)}y_{i}\right]\left\langle v_{j,L-1}\right\rangle \,,\label{eq: Delta L-1}
\end{eqnarray}
where we used a first order Taylor expansion, assuming $R_{ij,l}\ll1$.

Next, for $l<L$, we denote $c\left(i,l,l^{\prime}\right)$ to be
the index of the neuron in the $l^{\prime}$ layer which is receiving
input (through other neurons) from the $i$-th neuron in the $l$
layer, and $i_{m}=c\left(i,l,m\right)$ and $j_{m}=c\left(i,l,m-1\right)$.
We obtain

\begin{eqnarray}
R_{ij,l} & \approx & \ln\frac{\Phi\left(\frac{y_{i_{L}}\mu_{i_{L}\left(j_{L}\right),L}}{\sigma_{i_{L}\left(j_{L}\right),L}}\right)+\frac{1}{2}\tanh\left[\frac{2}{\sqrt{K_{L}}}\mathcal{N}\left(0|\mu_{i_{L}\left(j_{L}\right),L},\sigma_{i_{L}\left(j_{L}\right),L}^{2}\right)y_{i_{L}}\right]\left\langle W_{i_{L}j_{L},L}\right\rangle \left\langle v_{j_{L},L-1}|W_{ij,l}^{+}\right\rangle }{\Phi\left(\frac{y_{i_{L}}\mu_{i_{L}\left(j_{L}\right),L}}{\sigma_{i_{L}\left(j_{L}\right),L}}\right)+\frac{1}{2}\tanh\left[\frac{2}{\sqrt{K_{L}}}\mathcal{N}\left(0|\mu_{i_{L}\left(j_{L}\right),L},\sigma_{i_{L}\left(j_{L}\right),L}^{2}\right)y_{i_{L}}\right]\left\langle W_{i_{L}j_{L},L}\right\rangle \left\langle v_{j_{L},L-1}|W_{ij,l}^{-}\right\rangle }\nonumber \\
 & \approx & \tanh\left[\frac{2}{\sqrt{K_{L}}}\frac{\mathcal{N}\left(0|\mu_{i_{L}\left(j_{L}\right),L},\sigma_{i_{L}\left(j_{L}\right),L}^{2}\right)}{\Phi\left(y_{i_{L}}\mu_{i_{L}\left(j_{L}\right),L}/\sigma_{i_{L}\left(j_{L}\right),L}\right)}y_{i_{L}}\right]\left\langle W_{i_{L}j_{L},L}\right\rangle \left(\left\langle v_{j_{L},L-1}|W_{ij,l}^{+}\right\rangle -\left\langle v_{j_{L},L-1}|W_{ij,l}^{-}\right\rangle \right)\,,\label{eq: Delta L}
\end{eqnarray}
where we used a first order Taylor expansion, assuming $R_{ij,l}\ll1$
. We can now continue and apply Eq.~\ref{eq:P(v_m | x , W)} for
$m=L-1,L-2,\dots,l+1$, obtaining the recursive relation 
\begin{equation}
\frac{\left\langle v_{i{}_{m},m}|W_{ij,l}^{+}\right\rangle -\left\langle v_{i_{m},m}|W_{ij,l}^{-}\right\rangle }{\left\langle v_{i{}_{m-1},m-1}|W_{ij,l}^{+}\right\rangle -\left\langle v_{i_{m-1},m-1}|W_{ij,l}^{-}\right\rangle }\approx\tanh\left[\frac{2}{\sqrt{K_{m}}}\mathcal{N}\left(0|\mu_{i_{m}\left(j_{m}\right),m},\sigma_{i_{m}\left(j_{m}\right),m}^{2}\right)\right]\left\langle W_{i_{m}j_{m},m}\right\rangle \label{eq:delta m}
\end{equation}
We can now continue to apply Eq.~\ref{eq:P(v_m | x , W)}, until
we reach layer $l$, for which we need to use Eq.~\ref{eq: P(v_l | x W)}
to obtain

\begin{equation}
\left\langle v_{i,l}|W_{ij,l}^{+}\right\rangle -\left\langle v_{i,l}|W_{ij,l}^{-}\right\rangle \approx\tanh\left[\frac{2}{\sqrt{K_{l}}}\mathcal{N}\left(0|\mu_{i\left(j\right),l},\sigma_{i\left(j\right),l}^{2}\right)\right]\left\langle v_{j,l-1}\right\rangle \label{eq: delta l}
\end{equation}
or, if $l=1$

\begin{equation}
\left\langle v_{i,l}|W_{ij,l}^{+}\right\rangle -\left\langle v_{i,l}|W_{ij,l}^{-}\right\rangle \approx\tanh\left[\frac{2}{\sqrt{K_{l}}}\mathcal{N}\left(0|\mu_{i\left(j\right),l},\sigma_{i\left(j\right),l}^{2}\right)x_{j}\right]\,.\label{eq: delta l-1}
\end{equation}
Lastly, we note so far we assumed $R_{ij,l}\ll1$. This is not always
the case. For example, if $P\left(\mathbf{y}|W_{ij,l}^{-}\right)\rightarrow0$
and/or $P\left(\mathbf{y}|W_{ij,l}^{+}\right)\rightarrow0$, $R_{ij,l}$
may diverge. In this case the CLT theorem we used becomes invalid,
so we cannot use the expressions we previously found in section \ref{sec:Forward-propagation-of}
to obtain the correct asymptotic form in this limit - only the sign
of $R_{ij,l}$. From the conditions for the validity of the CLT we
can see qualitatively that this can occur if a highly ``unexpected''
sample $\left\{ \mathbf{x}^{\left(n\right)},\mathbf{y}^{\left(n\right)}\right\} $
arrive, when ``certainty'' of the posteriors increase - \emph{i.e.}
they converge either to zero or one. Unfortunately, as we stated,
the value of the log-likelihood (Eq.~\ref{eq: R_ij,l}) can diverge,
so the effects on the learning process can be quite destructive. As
an ad-hoc solution, in order to ward against the destructive effect
of such divergent errors, we just use the above expression even $R_{ij,l}\ll1$
is no longer true. In this case we will have some error in our estimate
of $R_{ij,l}$, but at least it will have the right sign and will
be finite in each time-step. Of course, even ``small'' errors can
sometimes accumulate significantly over long time, and we have no
mathematical guarantee this will not happen.

\section{Summary\label{sec:Summary}}

Next, we summarize our main results. This also appears in the paper,
but here we also give reference to the specific equations in the derivations.

In the  section \ref{sec:Forward-propagation-of} we defined (combining
the definitions in Eqs.~\ref{eq: mu for l=00003D1}, \ref{eq: sigma for l=00003D1},
\ref{eq: mu-1} and \ref{eq:sigma})
\begin{eqnarray*}
\mu_{k,m} & = & \frac{1}{\sqrt{K_{m}}}\sum_{r\in K\left(k,m\right)}\left\langle W_{kr,m}\right\rangle \left\langle v_{r,m-1}\right\rangle \\
\sigma_{k,m}^{2} & = & \frac{1}{K_{m}}\sum_{r\in K\left(k,m\right)}\left(\delta_{1m}\left(\left\langle v_{r,m-1}\right\rangle ^{2}-1\right)+1-\left\langle v_{r,m-1}\right\rangle ^{2}\left\langle W_{kr,m}\right\rangle ^{2}\right)
\end{eqnarray*}
and (combining the definitions in Eqs.~\ref{eq: mu 2}-\ref{eq: sigma 2}
with the two previous equations)
\begin{eqnarray*}
\mu_{i\left(j\right),l} & = & \mu_{i,l}-\frac{1}{\sqrt{K_{l}}}\left\langle W_{ij,l}\right\rangle \left\langle v_{j,l-1}\right\rangle \\
\sigma_{i\left(j\right),l}^{2} & = & \sigma_{i,l}^{2}-\frac{1}{K_{l}}\left(\delta_{1l}\left(\left\langle v_{j,l-1}\right\rangle ^{2}-1\right)+1-\left\langle v_{j,l-1}\right\rangle ^{2}\left\langle W_{ij,l}\right\rangle ^{2}\right)
\end{eqnarray*}
where (Eq.~\ref{eq: <v | x  >}) 
\[
\left\langle v_{k,m}\right\rangle =\begin{cases}
\mathbf{x} & ,\,\mathrm{if}\,\, m=0\\
2\Phi\left(\mu_{k,m}/\sigma_{k,m}\right)-1 & ,\,\mathrm{if}\,\, m>0
\end{cases}\,.
\]
Importantly, if we know $\mathbf{x}$ and $\left\langle W_{kr,m}\right\rangle =2P\left(W_{kr,m}^{+}\right)-1$
(recall $W^{\pm}$ means $W=\pm1$), all these quantities can be
calculated together in a sequential in ``forward pass'' for $m=1,2,...,L$. 

Using the above quantities, in section \ref{sec:The-log-likelihood-ratio}
we derived Eqs.~\ref{eq: Delta L}-\ref{eq: delta l}, which can
be summarized in the following concise way:
\begin{equation}
R_{ij,l}=\ln\frac{P\left(\mathbf{y}|W_{ij,l}^{+}\right)}{P\left(\mathbf{y}|W_{ij,l}^{-}\right)}=\begin{cases}
\Delta_{i^{\prime}i,l+1}\tanh\left[G_{ij,l}\right]\left\langle v_{j,l-1}\right\rangle  & ,\,\mathrm{if}\,\, l>1\\
\Delta_{i^{\prime}i,l+1}\tanh\left[G_{ij,l}x_{j}\right] & ,\,\mathrm{if}\,\, l=1
\end{cases}\label{eq: loglikelihood-1}
\end{equation}
where
\[
G_{ij,l}=\frac{2}{\sqrt{K_{l}}}\frac{\mathcal{N}\left(0|\mu_{i\left(j\right),l},\sigma_{i\left(j\right),l}^{2}\right)}{1+\left[\Phi\left(y_{i}\mu_{i\left(j\right),L}/\sigma_{i\left(j\right),L}\right)-1\right]\delta_{Ll}}
\]
and $\Delta_{ij,l}$ are defined recursively with $\Delta_{ij,L+1}=y_{j}$
and
\begin{equation}
\Delta_{ij,l}=\Delta_{i^{\prime}i,l+1}\left\langle W_{ij,l}\right\rangle \tanh\left[G_{ij,l}\right]\label{eq:Delta ij.l-1}
\end{equation}
with $i^{\prime}$ being the index of the neuron in the $m+1$ layer
receiving input from the $i$-th neuron in the $m$-th layer. Importantly,
all $\Delta_{ij,l}$ can be calculated in ``backward pass'' for
$m=L,\dots,2,1$. 

Now we can write down explicitly how $P\left(W_{ij,l}|D_{n}\right)$
change, according to the Bayes-based update rule in Eq.~$\ref{eq: bayes update-1 - L}$
\[
\ln\frac{P\left(W_{ij,l}^{-}|D_{n}\right)}{P\left(W_{ij,l}^{+}|D_{n}\right)}=\ln\frac{P\left(W_{ij,l}^{-}|D_{n-1}\right)}{P\left(W_{ij,l}^{+}|D_{n-1}\right)}+R_{ij,l}^{\left(n\right)}.\,
\]

\end{document}